\documentclass[lettersize,journal]{IEEEtran}
\usepackage{amsmath,amsfonts}
\usepackage{algorithm}
\usepackage{algpseudocode}
\usepackage{array}
\usepackage[caption=false,font=normalsize,labelfont=sf,textfont=sf]{subfig}
\usepackage{textcomp}
\usepackage{stfloats}
\usepackage{url}
\usepackage{verbatim}
\usepackage{graphicx}
\usepackage{cite}
\usepackage{hyperref}

\usepackage[capitalize,nameinlink]{cleveref}
\crefname{section}{Sec.}{Secs.}
\crefname{algorithm}{Alg.}{Algs.}
\crefname{appendix}{App.}{Apps.}
\crefname{definition}{Def.}{Defs.}
\crefname{table}{Table}{Tables}

\usepackage{xcolor}
\definecolor{vibrantred}{HTML}{E74C3C}
\definecolor{vibrantblue}{HTML}{3498DB}
\definecolor{vibrantgreen}{HTML}{2ECC71}
\definecolor{vibrantyellow}{HTML}{F1C40F}
\definecolor{vibrantpurple}{HTML}{9B59B6}
\definecolor{vibrantorange}{HTML}{E67E22}

\usepackage{booktabs} 
\usepackage{tabularray}
\usepackage{tabularx}
\usepackage{makecell}
\usepackage{wasysym}
\usepackage{multirow}

\usepackage[fontsize=0.02\paperwidth,angle=0,vpos=0.17\paperheight,firstpageonly=true]{draftwatermark}

\begin{document}

\title{Dexterous Robotic Piano Playing at Scale}

\author{Le Chen*, Yi Zhao*, Jan Schneider, Quankai Gao, Simon Guist, Cheng Qian, \\ Juho Kannala, Bernhard Schölkopf, Joni Pajarinen, and Dieter Büchler
\thanks{* Le Chen and Yi Zhao contributed equally to this work.}% <-this % stops a space
\thanks{L. Chen, J. Schneider, S. Guist, B. Schölkopf, and D. Büchler are with the Empirical Inference Department of Max Planck Institute for Intelligent Systems, 72076 Tübingen, Germany (e-mail: \{le.chen, jan.schneider, simon.guist, bernhard.schoelkopf, dieter.buechler\}@tuebingen.mpg.de).}
\thanks{Y. Zhao, J. Kannala, and J. Pajarinen are with Aalto University, 02150 Espoo, Finland (e-mail: \{yi.zhao, juho.kannala, joni.pajarinen\}@aalto.fi). }
\thanks{Q. Gao is with the University of Southern California, CA 90007, United States (e-mail: quankaig@usc.edu).}
\thanks{Q. Cheng is with Imperial College London, SW7 2AZ, London, United Kingdom (e-mail: c.qian24@imperial.ac.uk).}
\thanks{J. Kannala is with the University of Oulu, 90570 Oulu, Finland.}
\thanks{D. Büchler is also with the University of Alberta (Canada), the Alberta Machine Intelligence Institute~(Amii), \& holds a Canada CIFAR AI Chair.}
\thanks{Datasets, models, and code will be available upon acceptance.}
}

\SetWatermarkColor[rgb]{{1, 0.6, 0.6}}
\SetWatermarkText{This work has been submitted to the IEEE for possible publication. \vspace{-3pt} \\
Copyright may be transferred without notice, after which this version may no longer be accessible.}

\markboth{IEEE Transactions on Pattern Analysis and Machine Intelligence, in submission}%
{Shell \MakeLowercase{\textit{et al.}}: A Sample Article Using IEEEtran.cls for IEEE Journals}


\maketitle

\begin{abstract}
Endowing robot hands with human-level dexterity has been a long-standing goal in robotics.
Bimanual robotic piano playing represents a particularly challenging task: it is high-dimensional, contact-rich, and requires fast, precise control.
We present \textsc{OmniPianist}, the first agent capable of performing nearly one thousand music pieces via scalable, human-demonstration-free learning.
Our approach is built on three core components.
First, we introduce an automatic fingering strategy based on Optimal Transport (OT), allowing the agent to autonomously discover efficient piano-playing strategies from scratch without demonstrations.
Second, we conduct large-scale Reinforcement Learning~(RL) by training more than 2,000 agents, each specialized in distinct music pieces, and aggregate their experience into a dataset named RP1M++, consisting of over one million trajectories for robotic piano playing.
Finally, we employ a Flow Matching Transformer to leverage RP1M++ through large-scale imitation learning, resulting in the \textsc{OmniPianist} agent capable of performing a wide range of musical pieces.
Extensive experiments and ablation studies highlight the effectiveness and scalability of our approach, advancing dexterous robotic piano playing at scale.

\end{abstract}


\section{Introduction}

\begin{figure*}[t] 
\centering
\setlength{\abovecaptionskip}{0cm}
\includegraphics[width=0.99\textwidth]{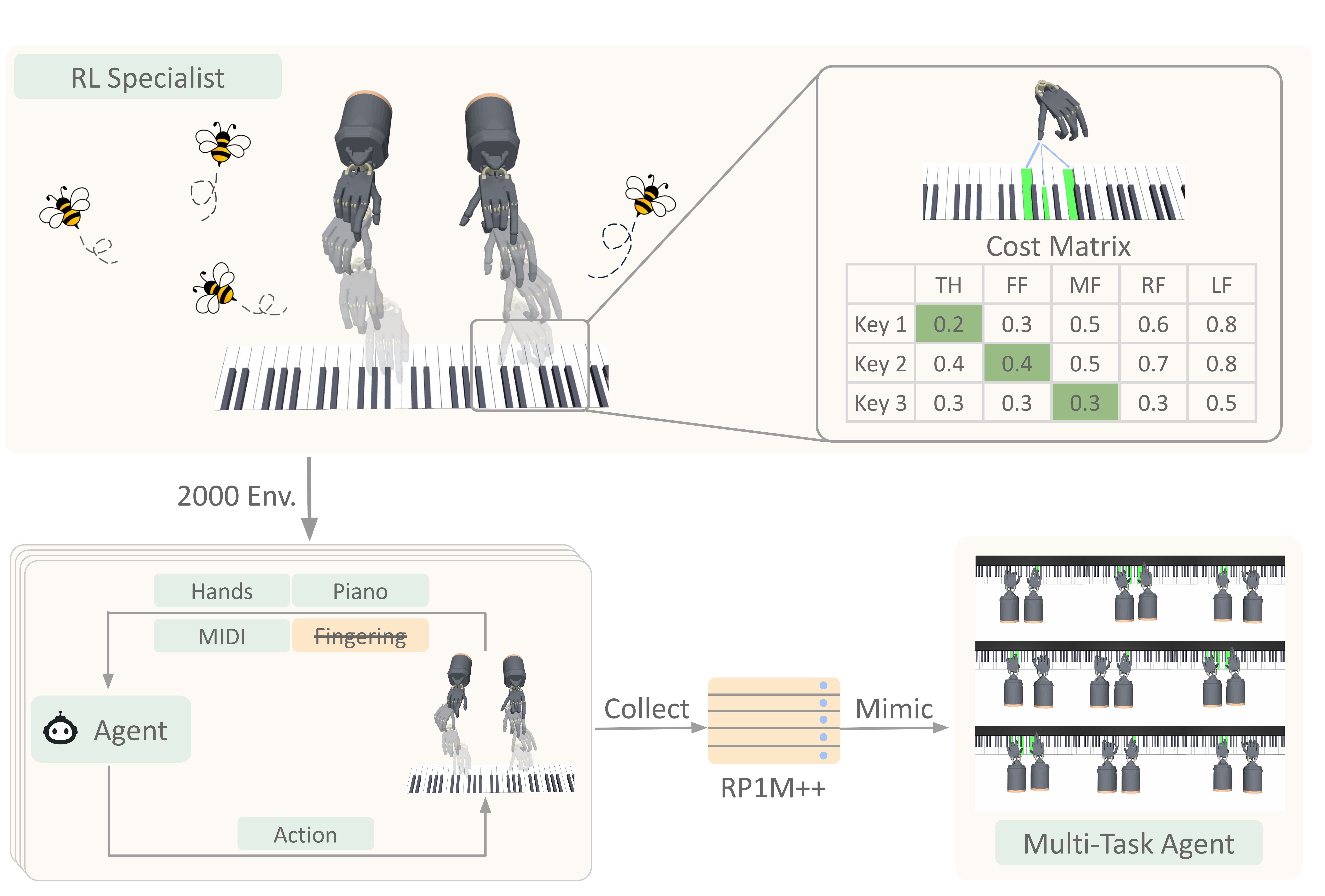}
\caption{Overview. This paper proposes an RL-based agent that alleviates the requirement for human-annotated fingering by formulating finger placement as an Optimal Transport (OT) problem, enabling the agent to play music pieces without human demonstrations, even for the challenging "Flight of the Bumblebee" song. We then collect a large-scale dataset named RP1M++ consists of more than 1 million expert piano-playing trajectories collected by training more than 2,000 RL agents. The collected data is consumed by a multi-task agent, named \textsc{OmniPianist}, via imitation learning that is capable of playing hundreds of musical pieces.}
\label{fig:overview}
\vspace{-0.4cm}
\end{figure*} 

Achieving human-level dexterity remains one of the central challenges in robotics.
The difficulty stems from the breadth of challenges ranging from contact-rich manipulation to dynamic athletic tasks, each posing distinct demands.
Manipulation tasks, such as grasping or reorienting objects~\cite{andrychowicz2020learning}, require sustained application of appropriate forces at moderate speeds across objects with diverse shapes, materials, and weight distributions.
Dynamic tasks, such as juggling~\cite{ploeger2021high} or table tennis~\cite{buchler2022learning}, involve frequent contact changes, demand high precision, and allow little tolerance for error due to the rarity of contact opportunities.
The high speeds in these settings necessitate large accelerations.
The combination of requiring both precision and speed makes reproducing human-level dexterity particularly challenging.
% The combination of requiring both precision and speed makes them particularly challenging.

Robot piano playing combines various aspects of both manipulation and dynamic tasks, making it a compelling yet underexplored testbed for dexterous control.
Like manipulation, it requires fine-grained coordination of multiple fingers to execute precise, timed contacts with small keys.
At the same time, rhythm, tempo, and musical expression introduce dynamic demands similar to athletic tasks.
Furthermore, while skilled pianists can perform arbitrary pieces with remarkable generalization, achieving such versatility remains far beyond current robotic capabilities.
In this work, we lay the foundation for robotic piano playing by developing methods that advance human-level bi-manual dexterity and explore generalization across diverse musical tasks.

While Reinforcement Learning (RL) is a promising approach for acquiring versatile skills in such challenging tasks, prior RL-based methods~\cite{zakka2023robopianist,qian2024pianomime,wang2024furelise} rely on costly human-annotated fingering information to solve piano playing.
In particular, the agent must know which finger should press each key at every time step to construct a dense reward signal.
This dependence on human annotations or demonstrations restricts learning to reproducing only labeled or demonstrated pieces, preventing the use of the vast repertoire of music available on the internet.
Unlike a human pianist, the agent trains based on a fixed fingering mapping that precludes improvising when, for instance, a hand is at an unusual position during play.
Moreover, such annotations may be impractical for robots with non-human morphologies, where variations in finger count, joint limits, or hand geometry make direct mapping from human fingering infeasible.

In this paper, we propose an approach that removes the need for human demonstrations by formulating finger placement as an Optimal Transport (OT) problem.
At each time step, our method solves an inner optimization problem: given the current fingertip positions and the keys to be pressed, it assigns fingers to keys so as to minimize overall movement distance.
This OT formulation provides a principled solution that directly guides hand movements for piano playing.
Our method matches the performance of human-annotated fingering with only a 3\% computation time overhead.
Moreover, because it makes no assumptions about hand morphology, it generalizes across different robotic hands and musical pieces, vastly expanding the range of songs available for training.

Another focus of this paper is enabling a single policy to play thousands of music pieces.
Although RL performs well in single-song settings, its performance degrades significantly when extended to multi-song scenarios~\cite{zakka2023robopianist}.
To address this, we first train specialist RL agents for individual pieces and then use their trajectories to train a multi-song agent via imitation learning.
However, prior results have shown that this approach suffers from clear performance drops as the dataset scales~\cite{zakka2023robopianist,zhao2024rp1m}.
A recent method, PianoMime~\cite{qian2024pianomime}, improves multi-song imitation learning through a hierarchical policy combined with inverse kinematics, but at the cost of a complex and difficult-to-scale training pipeline.

Through careful analysis, we identify two primary factors behind the poor performance of previous multi-task training~\cite{zhao2024rp1m}: limited data diversity and insufficient model expressiveness.
Although the previous RP1M dataset~\cite{zhao2024rp1m} contains over one million trajectories from more than 2,000 RL agents, these trajectories are generated by unrolling expert policies.
As a result, the data covers only a narrow region of the state space, leading to performance degradation when the agent deviates from this manifold.
To address this limitation, we construct a new dataset, RP1M++, by collecting data generated when training agents from scratch with Dataset Aggregation (DAgger)~\cite{ross2011reduction}, thereby collecting trajectories with substantially higher state diversity.
RP1M++ significantly improves imitation learning performance by broadening the coverage of training data.
Beyond data, we also leverage a Flow Matching Transformer~\cite{lipman2022flow,esser2024scaling}, which delivers notable gains over Diffusion Models~\cite{chi2023diffusion}.
With improvements from both the dataset and the policy representation, our multi-song agent, \textsc{OmniPianist}, sustains high performance as the dataset scales and demonstrates strong generalization to unseen songs, achieving an average F1 score of 0.55 on 100 novel songs.

To summarize, our paper makes the following contributions:
\begin{itemize}
\item We introduce an automatic fingering placement strategy based on OT, enabling RL agents to play piano without any form of human demonstration.
\item We conduct large-scale RL by training over 2,000 specialist agents, each dedicated to a distinct music piece, and use them to collect RP1M++, a dataset of more than one million diverse trajectories for robotic piano playing.
\item We propose \textsc{OmniPianist}, a Flow Matching Transformer that masters a wide range of musical pieces and demonstrates promising generalization to unseen songs.
\end{itemize}

This paper extends our previous conference paper RP1M~\cite{zhao2024rp1m} with the following new contributions:
\begin{itemize}
    \item We analyze the key failure reasons of training a multi-song robotic piano-playing agent and propose fixes that enable scalable training with strong performance.
    \item We construct RP1M++, a more diverse dataset with broader state-space coverage, which is key to \textsc{OmniPianist}’s success.
    \item We validate our OT-based fingering strategy on additional embodiments, including the Allegro~\cite{allegro} and ORCA~\cite{christoph2025orca} hands.
    \item We provide extensive ablation studies analyzing the impact of the proposed RP1M++ dataset and the Flow Matching Transformer.
\end{itemize}

\section{Related Work}

\newcolumntype{Y}{>{\centering\arraybackslash}X}
\begin{table*}[t]
    \caption{Existing datasets on dexterous or bimanual robotic manipulation.}
    \label{tab:existing_robot_datasets}
    \centering
    \footnotesize
    \small
    \setlength\tabcolsep{5pt}
    \begin{tabularx}{\textwidth}{cYYYYY}
        \specialrule{1pt}{1pt}{2.5pt}
        Dataset & Task & \makecell{Dexterous \\ hands} & Bimanual & \makecell{Dynamic \\ tasks} & \makecell{Demonstrations} \\
        \specialrule{1pt}{1pt}{2.5pt}
        DexGraspNet~\cite{wang2023dexgraspnet} & grasping & \textbf{\checked} & & & 1.3M \\
        \midrule
        RealDex~\cite{liu2024realdex} & grasping & \textbf{\checked} & & & 2.6K\\
        \midrule
        UniDexGrasp~\cite{xu2023unidexgrasp} & grasping & \textbf{\checked} & & & 1.1M \\
        \midrule
        ALOHA~\cite{zhao2023learning} & manipulation & & \textbf{\checked} & & 825 \\
        \midrule
        Bi-DexHands~\cite{chen2023bi} & manipulation & \textbf{\checked} & \textbf{\checked} & partially & $\sim$20K \\ 
        \midrule
        D4RL~\cite{fu2020d4rl} (Adroit) & manipulation & \textbf{\checked} & & & 30K\\
        \specialrule{1pt}{1pt}{2.5pt}
        RP1M++ (ours) & piano & \textbf{\checked} & \textbf{\checked} & \textbf{\checked} & $\sim$1M \\
        \specialrule{1pt}{1pt}{2pt}
    \end{tabularx}
    % \vspace{-0.1cm}
\end{table*}

\textbf{Dexterous Robot Hands}~~
Research on dexterous robot hands seeks to replicate human hand dexterity in robotic systems.
Model-based approaches~\cite{rus1999hand,bicchi1995dexterous,han1998dextrous,bai2014dexterous,mordatch2012contact,dafle2014extrinsic,chavan2020sampling,kumar2014real} employ planning algorithms to compute trajectories executed by controllers, requiring accurate robot hand models.
While closed-loop methods incorporating sensor feedback~\cite{li2014learning} have been developed, they similarly demand precise hand models, which is particularly challenging given the complex multi-contact dynamics between hands and objects.

Due to the difficulty of accurately modeling dexterous hand dynamics, recent research has shifted toward learning-based approaches, particularly RL~\cite{akkaya2019solving,andrychowicz2020learning}. 
To address the training complexity associated with high-dimensional action spaces, demonstrations are commonly used~\cite{kumar2016learning,rajeswaran2017learning,radosavovicstate,jeong2020learning,lin2024learning}.
Recent advances in RL algorithms and simulation have enabled impressive manipulation performance without human demonstrations, with policies successfully transferred from simulation to real dexterous hands~\cite{chen2022system,yang2022learning,chen2022towards,xu2023unidexgrasp,allshire2022transferring,qin2023dexpoint,lin2024twisting}.

\textbf{Piano Playing with Robots}~~
Robotic piano playing has been investigated for decades, presenting significant challenges due to the precise bimanual coordination required to press correct keys with accurate timing in high-dimensional action spaces. 
Early approaches relied on specialized robot designs~\cite{kato1987robot,lin2010electronic,hughes2018anthropomorphic,castro2022robotic,zhang2009design} or pre-programmed trajectories~\cite{li2013controller,zhang2011musical}. 
Recent methods have enabled piano playing with dexterous hands through planning~\cite{scholz2019playing} or RL~(in simulation~\cite{xu2022towards} and using physical robot hands~\cite{zeulner2025learning}), though performance remains limited to simple musical pieces.

RoboPianist~\cite{zakka2023robopianist} introduced a comprehensive benchmark for robotic piano playing, demonstrating strong RL performance while requiring human fingering annotations and exhibiting degraded multi-task learning capabilities. 
Human fingering annotations specify finger-key correspondences at each timestep labeled by human experts, making them expensive to obtain~\cite{wang2024furelise}. 
Various approaches have learned fingering from human-annotated data using different machine learning methods~\cite{nakamura2020statistical,ramoneda2022automatic,randolph2023modeling}.

To reduce annotation costs, several works have explored automatic fingering acquisition.
\cite{moryossef2023your} extract fingering from videos, while~\cite{ramoneda2021piano}
formulates piano fingering as sequential decision-making using RL but without considering robot hand modeling.
\cite{shi2022optimized} automatically acquires fingering via dynamic programming, but the solution is limited to simple tasks. 
Concurrent work~\cite{qian2024pianomime} derives fingering from YouTube videos through human motion extraction and retargeting.
In contrast to these approaches requiring explicit fingering models, our method enables fingering to emerge automatically during piano playing, similar to how humans learn to play piano, thereby expanding the available training data for training multi-task piano-playing agents.

\textbf{Datasets for Dexterous Robot Hands}~~
Most existing large-scale dexterous robot hand datasets focus on object grasping tasks. 
These datasets employ various approaches to determine grasp configurations, e.g., planning-based methods~\cite{liu2020deep,wang2023dexgraspnet,murrilo2024multigrippergrasp}, learned grasping policies~\cite{xu2023unidexgrasp}, or human motion tracking and imitation~\cite{liu2024realdex}.

In contrast to the abundance of grasping datasets, few datasets exist for dexterous manipulation tasks. 
The D4RL benchmark~\cite{fu2020d4rl} provides small sets of expert trajectories for four manipulation tasks, consisting of human demonstrations and rollouts of trained policies. 
\cite{zhao2023learning} presents a small object manipulation dataset that utilizes a low-cost bimanual platform with simple parallel grippers.
\cite{chen2023bi} collects offline datasets for two simulated bimanual manipulation tasks with dexterous hands. 
Furthermore, ARCTIC~\cite{fan2023arctic} proposes a large-scale dataset for bimanual hand-object manipulation, though focused on human rather than robot hands.
To the best of our knowledge, the RP1M dataset~\cite{zhao2024rp1m} is the first large-scale dataset of dynamic, bimanual piano playing with dexterous robot hands.
In this paper, we extend RP1M with a much diverse data distribution, named RP1M++. As shown later, the diverse dataset is the key to the success of training multi-task agents.

\textbf{Robot Learning from Demonstration}~~
While RL methods often perform well on single tasks, humans excel at executing diverse tasks in dynamic environments. To bridge this gap, multi-task agents have been developed to learn a wide range of skills within a unified model~\cite{reed2022generalist,lee2020learning,brohan2022rt,bharadhwaj2024roboagent}. These approaches typically rely on scalable architectures and large-scale datasets to generalize across tasks~\cite{brohan2022rt,brohan2023rt,ha2023scaling,du2024learning,bonatti2022pact}.

Diffusion models have recently emerged as a powerful generative modeling approach, achieving state-of-the-art results across various generative tasks, including image~\cite{rombach2022high}, video~\cite{ho2022video,ho2022imagen}, and 3D content generation~\cite{poole2022dreamfusion,liu2023meshdiffusion}. In robotics, diffusion models have been employed as policy networks for imitation learning in domains such as manipulation~\cite{chi2023diffusion,ha2023scaling,reuss2023goal,ghosh2024octo}, navigation~\cite{sridhar2024nomad}, and locomotion~\cite{huang2024diffuseloco}, and have also been applied to multi-task learning settings~\cite{ha2023scaling,ghosh2024octo,barreiros2025careful}. However, a major limitation of diffusion models is their slow inference speed due to the requirement of many iterative denoising steps.

Flow matching offers a promising alternative. Rooted in Optimal Transport theory, flow matching learns a vector field that defines an ordinary differential equation whose solution traces the desired probabilistic path between a base distribution and the target data distribution~\cite{lipman2022flow}. Compared to diffusion models, flow matching is more straightforward, requiring fewer hyperparameters, and exhibits greater numerical stability. Flow matching has demonstrated superior performance over diffusion models in various domains, including image generation~\cite{esser2024scaling}, protein structure modeling~\cite{bose2023se}, video synthesis~\cite{jin2024pyramidal}, and robotic manipulation~\cite{hu2024adaflow,chisari2024learning,zhang2024affordance}. In the context of robotics, flow matching has shown great potential in modeling complex, multimodal action distribution and serving as an expressive policy representation for imitation learning~\cite{black2410pi0}. 
In this work, we investigate the application of flow matching in high-dimensional control tasks, that is, playing piano with bimanual dexterous robot hands.

\section{Preliminaries}
\label{sec:pre}

\textbf{Reinforcement Learning}~~ In Reinforcement Learning~(RL), an agent or policy $\pi(a|s)$ updates its parameters using trajectories $\tau = \{s_t, a_t, r_t, s_{t+1}\}_{t=0}^T$ collected from the environment, where $s_t$, $a_t$, $r_t$ are states, actions and rewards at time step $t$, respectively. 
Formally, RL solves a sequential decision-making problem formulated as a Markov Decision Process (MDP).
An MDP can be defined as $<\mathcal{S}, \mathcal{A}, \mathcal{T}, \mathcal{R}, \gamma>$, where $\mathcal{S}$ denotes the set of states, $\mathcal{A}$ the set of actions, $\mathcal{T}$ the state-transition dynamics, $\mathcal{R}$ the reward function, and $\gamma$ the discount factor.
The goal of RL is to learn a policy $\pi(a|s)$ that outputs actions to maximize the expected returns:
\begin{equation} \label{eq:background/obj}
    J(\theta) = \mathbb{E}_{p_\theta(\tau)} \Big[ \sum_{t=0}^T \gamma^t r_t\Big] = \mathbb{E}_{p_\theta(\tau)} \Big[ R(\tau) \Big].
\end{equation}

To optimize the objective in~\cref{eq:background/obj}, we first learn an optimal action-value function $Q(s,a)$ to approximate the best possible long-term return achievable from state $s$ taking action $a$. The training target is given as $y= r+\gamma Q(s',\tilde{a})$, $\tilde{a} \sim \pi_\theta(\cdot|s')$. To stabilize the training and encourage exploration, we deploy the training target used in~\cite{hiraoka2021dropout} by taking the minimal value over multiple Q functions and introducing a policy entropy term. For transitions $(s,a,r,s')$ sampled from replay buffer $\mathcal{D}$, the training target used in this paper is:
\begin{equation}\label{eq:background/q_target}
    y = r + \gamma \Big( \min_{i=1,\cdots,M} Q_{\phi_i}(s',\tilde{a}) - \alpha \log \pi_\theta(\tilde{a}|s') \Big),   ~~\tilde{a}\sim\pi_\theta(\cdot | s').
\end{equation}
The policy $\pi_\theta(a|s)$ is optimized by maximizing the learned Q function while maximizing the policy entropy for better exploration. The training objective is given as:
\begin{equation} \label{eq:background/policy}
    \mathcal{L}(\theta) = -\mathbb{E}_{s\sim\mathcal{D}}\Big[Q(s, \tilde{a}) - \alpha \log \pi_\theta (\tilde{a}|s)\Big], ~~\tilde{a} \sim \pi_\theta(\cdot|s).
\end{equation}

\textbf{Optimal Transport}~~
Optimal Transport (OT), initially investigated by Monge, finds the most cost-efficient plan to move one "pile of mass" into another~\cite{peyre2019computational}. 
Given a source distribution $\mu$ and a target distribution $\nu$, OT aims to find a map $T:X \rightarrow Y$ such that:
\begin{equation} \label{eq:background/ot}
    \min_{T_{\#}\mu=\nu} \int c(x, T(x))~d\mu(x).
\end{equation}
$T_{\#}\mu=\nu$ represents a map $T$ that pushes $\mu$ to $\nu$, $c(x,y)$ is a cost function of transporting $x$ to $y$.
In this paper, we use Euclidean metric for the cost function, i.e., $c(x,y) = ||x-y||_2$.
The solution of solving the optimization problem from~\cref{eq:background/ot} gives an optimal plan to do the transportation as well as the minimal transporting cost $d$. We refer to~\cite{peyre2019computational} for a detailed introduction. 

\textbf{Diffusion Policy}~~
Diffusion models, such as Denoising Diffusion Probabilistic Models (DDPM)~\cite{ho2020denoising}, are generative models that produce data samples via a learned denoising process~\cite{welling2011bayesian, sohl2015deep}. They define a Markov chain consisting of diffusion steps that progressively introduce Gaussian noise to the original data, allowing the model to learn how to reverse this process and reconstruct desired samples from pure noise. Specifically, given a continuous-valued data distribution $p(x^{0})$, the forward diffusion process iteratively adds Gaussian noise $\epsilon \sim \mathcal{N}(0, \mathbf{I})$ to the data:
\begin{equation}
q(x^{k}\mid x^{k-1}) = \mathcal{N}\left(x^{k}; \sqrt{1 - \beta_k}\,x^{k-1}, \beta_k\,\mathbf{I}\right),
\end{equation}
where $\beta_k$ denotes the variance schedule at diffusion step $k$.
The reverse process performs denoising, guided by a neural network parameterized by $\theta$ that predicts the noise $\epsilon_{\theta}(x^k,k)$ added during the forward steps. Sampling begins from pure Gaussian noise $x^{K}\sim \mathcal{N}(0, \mathbf{I})$ and iteratively removes noise according to:
\begin{equation}
p_\theta(x^{k-1}\mid x^k) = \mathcal{N}\left(x^{k-1}; \mu_{k}(x^{k}, \epsilon_{\theta}(x^k,k)), \sigma_k^2\,\mathbf{I}\right),
\end{equation}
where $\mu_k$ is a predefined function that computes the mean of the reverse distribution at step $k$, based on the current noisy sample $x^k$ and the predicted noise $\epsilon_\theta$. This iterative denoising procedure enables the reconstruction of target data samples from initial noise.

Diffusion Policy~\cite{chi2023diffusion} is a diffusion-based imitation learning framework that has demonstrated strong performance across a range of simulated and real-world robotic tasks. The policy $\pi_\theta$ leverages the DDPM objective to learn a generative model over action sequences $a$, conditioned on observations $s$. It frames action generation as a denoising process $p_\theta(a^{k-1}|a^k, s)$ and can be trained via behavior cloning by fitting the conditional noise prediction $\epsilon_{\theta}(a^k, s, k)$. In this way, Diffusion Policy effectively models complex, multimodal behavior distributions, making it particularly well-suited for high-dimensional control tasks, such as robotic piano playing, where generating diverse, precise, and temporally coherent actions is essential. Diffusion is closely related to the flow-matching framework we use and serves as an important baseline in our experiments.

\section{Training Specialist Piano-Playing Agents}
\label{sec:specialist}
In this section, we present our approach for training specialist agents, each dedicated to mastering a specific piano piece using bi-manual dexterous robotic hands.
We formulate the piano-playing task as a finite-horizon MDP and train the agents using RL.
The task setup, including the observation space, action space, and evaluation metrics, is described in \cref{specialist:task}.
An OT-based fingering placement strategy is introduced in \cref{specialist:ot}, which allows agents to learn piano playing without demonstrations.
Finally, the overall reward function is detailed in \cref{specialist:reward}.

\subsection{Task setup}
\label{specialist:task}
The simulated piano-playing environment builds upon RoboPianist~\cite{zakka2023robopianist}. It consists of three main components: a robotic piano-playing setup, an RL-based agent for piano playing using simulated robot hands, and a multi-task learning framework. For clarity, we refer to these components as \textit{RoboPianist}, \textit{RoboPianist-RL}, and \textit{RoboPianist-MT}, respectively. The piano playing environment features a full-sized keyboard with 88 keys driven by linear springs, two Shadow robot hands~\cite{shadow-robot}, and a simulated pseudo sustain pedal.
Musical scores are represented using Musical Instrument Digital Interface (MIDI) transcriptions. Each time step in the MIDI file specifies which piano keys should be pressed (i.e., active keys). The objective of the piano-playing agent is to accurately press these active keys while avoiding inactive ones, under both spatial and temporal constraints. This task requires the agent to coordinate its fingers and place them properly in a highly dynamic scenario, such that target keys, not only at the current time step but also at future time steps, can be pressed accurately and in a timely manner. The original RoboPianist leverages MIDI files from the PIG dataset~\cite{nakamura2020statistical}, which include expert-annotated \textit{human fingering} information. However, as previously discussed, this reliance restricts the agent to playing only human-labeled music pieces, and the human annotation may not be suitable for robots due to its fixed nature and differences in morphologies. 

The observation includes the state of two robot hands, fingertip positions, piano sustain state, piano key states, and a goal vector. The goal vector consists of the current step's goal and the goals of multiple future steps. At each time step, an 89-dimensional binary vector is used to represent the goal, where 88 dimensions correspond to key states and the last dimension corresponds to the sustain pedal. In this paper, we include 10 future steps of active keys and sustain pedal targets, resulting in a 1144-dimensional observation space. While the original RoboPianist further includes 10-step human fingering labels as part of the observation, we omit this information in our setup, as our approach does not depend on human-labeled fingering. To define the action space, we exclude degrees of freedom that are not present in the human hand or are rarely used in typical piano performances. This results in a 39-dimensional action space comprising the joint positions of the robot hands, the positions of the forearms, and the control of the sustain pedal.

We assess the performance of the trained agent using the average F1 score, defined as $F_1 = 2 \cdot \frac{\text{precision} \cdot \text{recall}}{\text{precision} + \text{recall}}$. In this context, recall measures the agent’s success in pressing the correct (active) keys, while precision reflects its ability to avoid pressing incorrect (inactive) ones~\cite{zakka2023robopianist}.

\begin{algorithm}[t]
\caption{Training RL Specialist Agents for Piano Playing}\label{alg:rl}
\begin{algorithmic}
\Require Randomly initialized policy $\pi_\theta$, Q functions $Q_{\phi_i}$, $i=1,\dots M$. 
        Set target Q parameters as $Q_{\bar{\phi}_i} \leftarrow Q_{\phi_i}$.
        Replay buffer $\mathcal{D}_{\text{}} \gets \emptyset$. 
\For {training episode $N$}
    \State \textcolor{vibrantblue}{\emph{// Collect Trajectories}}
    \State $t \leftarrow 0$
    \While{$t \leq \text{episode length}~H$}
        \State Obtain action $a_t \sim \pi_{\theta}(a_t|o_t)$ with the policy.
        \State Interact the env. with $a_t$ to obtain $(o_t, a_t,r_t,o_{t+1})$, where $r_t$ is calculated with~\cref{eq:rl/overall_reward}
        \State Store transition $\mathcal{D} \gets (o_t, a_t, r_t,o_{t+1})$
    \EndWhile

    \State \textcolor{vibrantblue}{\emph{// Update policy and Q functions}}
    \For {$G$ updates}
        \State Sample a batch $\mathcal{B}=\{o_i,a_i,r_i,o_i'\}^{|\mathcal{B}|}_{i=0}$ from $\mathcal{D}$.
        \State \textcolor{vibrantorange}{\emph{// Update Q functions}}
        \State Compute the Q target $Q_\text{tar}$ according to~\cref{eq:background/q_target}.
        \For{$i=1,\dots,M$}
            \State Update $\phi_i$ by minimizing \\ ~~~~~~~~~~~~~~~~~~~~~~$\mathcal{L}(\phi_i) = \frac{1}{|\mathcal{B}|}\sum_{(o,a,r,o')\in \mathcal{B}}(Q_{\phi_i}(o,a)-Q_{\text{tar}})^2$.
            \State Update target networks $\bar{\phi_i} \gets \rho \bar{\phi_i}  + (1-\rho)\phi_i$.
        \EndFor
        \State \textcolor{vibrantorange}{\emph{// Update policy $\pi_\theta$}}
    \State Update $\theta$ by minimizing~\cref{eq:background/policy}.
    \EndFor
    
\EndFor

\end{algorithmic}
\end{algorithm}

\subsection{Automatic Fingering Annotation with Optimal Transport}
\label{specialist:ot}
To mitigate the hard exploration problem posed by the sparse rewards, \textit{RoboPianist-RL}~\cite{zakka2023robopianist} adds dense reward signals by using human fingering labels. Fingering informs the agent of the ``ground-truth" fingertip positions, and the agent minimizes the Euclidean distance between the current fingertip positions and the ``ground-truth" positions. 
However, acquiring such fine-grained annotations from expert human pianists is both time-consuming and expensive. A recent work~\cite{qian2024pianomime} addresses this by replacing manual fingering labels with hand motion trajectories extracted from YouTube videos of human piano performances, enabling agents to learn from passive visual demonstrations. Nevertheless, approaches that rely on human annotations or demonstrations inherently limit the agent's ability to generalize, as they constrain training to music pieces that already have labeled fingerings or available performance videos. In practice, a vast number of songs lack such resources, making it difficult to scale learning using these methods alone.
In this paper, we propose an automatic fingering annotation method based on OT to lift the requirement of human-annotated fingering or demonstrations.

Although fingering is highly personalized, it generally serves to enable pianists to press keys in a timely and efficient manner. Inspired by this principle, we not only aim to maximize rewards associated with correct key presses but also seek to minimize the movement distance of the fingers.  Specifically, at time step $t$, for the $i$-th key $k^i$ to be pressed, we assign the $j$-th finger $f^j$ to press this key, such that the total finger movement cost is minimized. We define the minimized cumulative moving distance between fingers and target keys as $d_t^{\text{OT}} \in \mathbb{R}^+$, computed as follows:
\begin{equation} \label{eq:ot}
\begin{aligned}
d_t^{\text{OT}} = \min_{w_t} \sum_{(i,j)\in K_t \times F} w_t(k^i, f^j)\cdot c_t(k^i, f^j) \\
\text{s.t.}~ \textcolor{vibrantblue}{\underbrace{\normalcolor{\sum_{j\in F} w_t(k^i, f^j) = 1,~ \forall i \in K_t}}_{\scriptstyle{\emph{\text{each key pressed by one finger only}}}}}, \\
\quad~ \textcolor{vibrantblue}{\underbrace{\normalcolor{\sum_{i\in K_t} w_t(k^i, f^j) \leq 1,~ \forall j \in F}}_{\scriptstyle{\emph{\text{each finger can press at most one key}}}}}, \\
\quad~ \textcolor{vibrantblue}{\underbrace{\normalcolor{w_t(k^i, f^j) \in \{0, 1\},~ \forall (i, j) \in K_t \times F}}_{w_t(k^i, f^j)\scriptstyle{\emph{ \text{is binary}}}}}.
\end{aligned}
\end{equation}

$K_t$ represents the set of keys to be pressed at time step $t$, and $F$ represents the set of available fingers on the robot hands.
The cost function $c_t(k^i, f^j)$ represents the Euclidean distance between finger $f^j$ and piano key $k^i$ at time step $t$, capturing the effort required to move the finger to the target key. The assignment variable $w_t(k^i, f^j)$ is a binary indicator that specifies whether finger $f^j$ is assigned to press key $k^i$. In our formulation, each key in $K_t$ must be pressed by exactly \textit{one} finger in $F$, and each finger can press \textit{at most} one key. 
The constrained optimization problem in Eq.~\ref{eq:ot} is a classic instance of an OT problem. Intuitively, it tries to find the best ``transport" strategy such that the overall cost of moving (a subset of) fingers $F$ to keys $K_t$ is minimized.
We solve this optimization problem with a modified Jonker-Volgenant algorithm~\cite{crouse2016implementing,virtanen2020scipy}. The resulting optimal assignments $(i^*, j^*)$ determine the finger-to-key pairings, which we treat as the agent’s fingering decisions. This fingering is computed \emph{dynamically at each time step} based on the current state of the robot hands, allowing it to adapt continuously during RL training. As a result, the agent’s fingering strategy evolves in response to its morphology and motion, enabling more efficient and physically consistent piano playing.
We define a reward $r_t^{\text{OT}}$  to encourage the agent to move the fingers close to the keys $K_t$. which is defined as:
\begin{equation}
    r_t^{\text{OT}} = 
    \begin{cases}
    \exp(c\cdot (d_t^{\text{OT}}-\delta)^2) ~~~&\text{if} ~~d_t^{\text{OT}} \geq \delta,
    \\
    1.0 ~~~&\text{if} ~~d_t^{\text{OT}} < \delta.
    \end{cases} \label{eq:r_ot}
\end{equation}
The constant $c$ is a negative scaling factor, following the setting used in~\cite{tassa2018deepmind}, and $d_t^{\text{OT}}$ represents the OT distance between fingers and target keys, obtained by solving Eq.~\ref{eq:ot}. $r_t^{\text{OT}}$ increases exponentially as $d_t^{\text{OT}}$ decreases and is set as 1 once $d_t^{\text{OT}}$ is smaller than a pre-defined threshold $\delta$, which is set to 0.01 in our case.

\subsection{Reward Function}
\label{specialist:reward}
The overall reward function is defined as:
\begin{equation} \label{eq:rl/overall_reward}
    r_t = r^{\text{OT}}_t + r^{\text{Press}}_t + r^{\text{Sustain}}_t + \alpha_1 \cdot r^{\text{Collision}}_t + \alpha_2 \cdot r^{\text{Energy}}_t
\end{equation}
$r_t^{\text{Press}}$ indicates whether the active keys are correctly pressed and inactive keys are not pressed. We use the same implementation as~\cite{zakka2023robopianist}, given as:
$r_t^{\text{Press}} = 0.5 \cdot (\frac{1}{K} \sum_{i=1}^K g(||k_t^i - 1||_2)) + 0.5 \cdot (1 - \textbf{1}_{\text{false positive}})$. K is the number of active keys, $k_t^i$ is the normalized key state of key $i$ at time step $t$ with range [0, 1], where 0 means the $i$-th key is not pressed and 1 means the key is pressed. $g(\cdot)$ is a tolerance function from DeepMind Control Suite~\cite{tassa2018deepmind}, which is similar to the one used in~\cref{eq:r_ot}. $\mathbf{1}_{\text{false positive}}$ indicates whether the inactive keys are pressed, which encourages the agent to avoid pressing keys that should not be pressed. 
$r_t^{\text{Sustain}}$ encourages the agent to press the pseudo sustain pedal at the right time, given as $r_t^{\text{Sustain}} = g(s_t - s_t^{\text{target}})$. $s_t$ and $s_t^{\text{target}}$ are the states of the current and target sustain pedal, respectively. 
$r_t^{\text{Collision}}$ penalizes the agent from collision, defined as $r_t^{\text{Collision}} = 1 -\textbf{1}_\text{collision}$, where $\textbf{1}_\text{collision}$ is 1 if collision happens and 0 otherwise.
$r_t^{\text{Energy}}$ prioritizes energy-saving behavior. 
It is defined as $r_t^{\text{Energy}}=-|\tau_{\text{joints}}|^\top |\mathbf{v}_{\text{joints}}|$. $\tau_{\text{joints}}$ and $\mathbf{v}_{\text{joints}}$ are joint torques and joint velocities respectively.
$\alpha_1$ and $\alpha_2$ are coefficient terms, and $\alpha_1=0.5$ and $\alpha_2=5\cdot 10^{-3}$ are adopted. 

Our method is compatible with any RL method, and we use DroQ~\cite{hiraoka2021dropout} in our paper. DroQ is a model-free RL method that uses Dropout~\cite{srivastava2014dropout} and Layer Normalization~\cite{ba2016layer} in the Q function to improve sample efficiency. The pseudocode is given in~\cref{alg:rl}. 
During RL training, the fingering is computed on-the-fly by solving~\cref{eq:ot}, continuously updating to best match the current hand configuration and fingertip positions.

\section{Large-Scale Motion Dataset Collection}
\label{sec:dataset}
By removing the dependence on human-provided fingering labels, our approach enables agents to learn from any sheet music available online (subject to copyright permissions), significantly expanding the range of playable pieces. To further support research on dexterous robotic hands, we collect and publicly release a large-scale piano-playing motion dataset, providing a rich resource for training and benchmarking future models.

\subsection{RP1M Dataset}
In the conference version of this paper~\cite{zhao2024rp1m}, we collect \textit{RP1M}, a dataset of piano playing motions that includes more than 2k music pieces with expert trajectories generated by our agents. It includes $\sim$1M expert trajectories covering $\sim$2k musical pieces. 
For each musical piece, we train an individual DroQ agent with the OT-based method introduced previously for 8 million environment steps and collect 500 expert trajectories by rolling out the trained policy with different seeds and noises. 
We chunk each sheet of music every 550 time steps, corresponding to 27.5 seconds, so that each run has the same episode length. 
The sheet music used for training is from the PIG dataset~\cite{nakamura2020statistical} and a subset (1788 pieces) of the GiantMIDI-Piano dataset~\cite{kong2020giantmidi}. 

\begin{figure*}[t] 
\setlength{\abovecaptionskip}{0cm}
\centering
\includegraphics[width=0.99\textwidth]{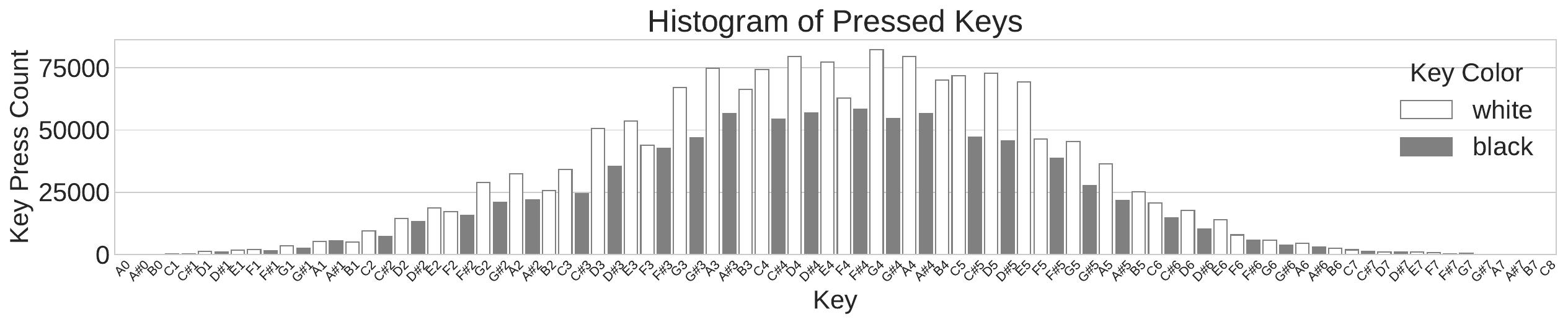}
\includegraphics[width=0.47\textwidth]{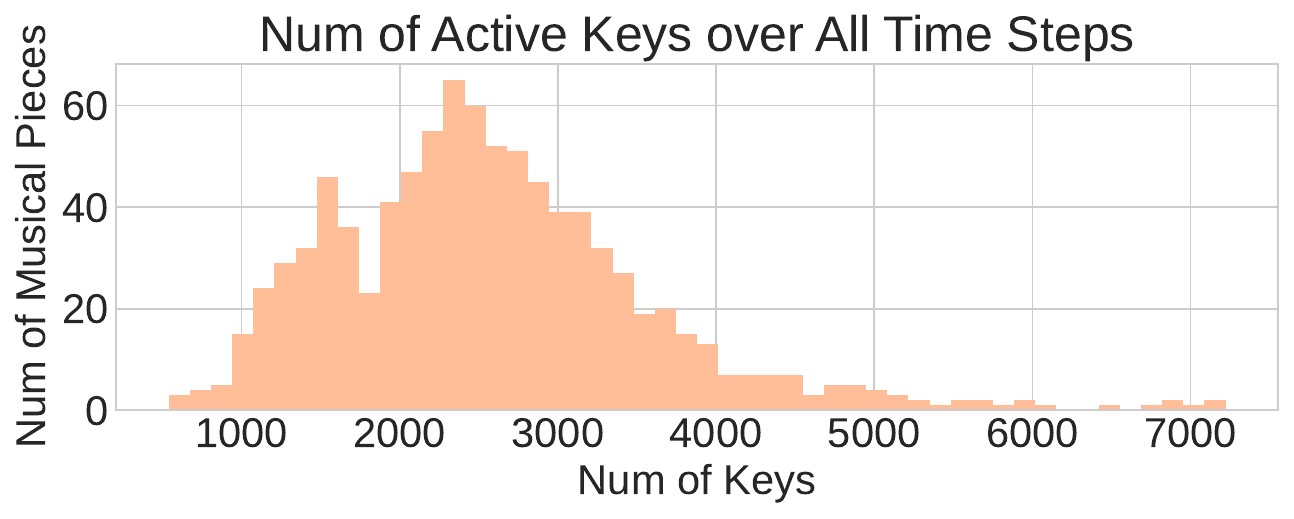}
\includegraphics[width=0.52\textwidth]{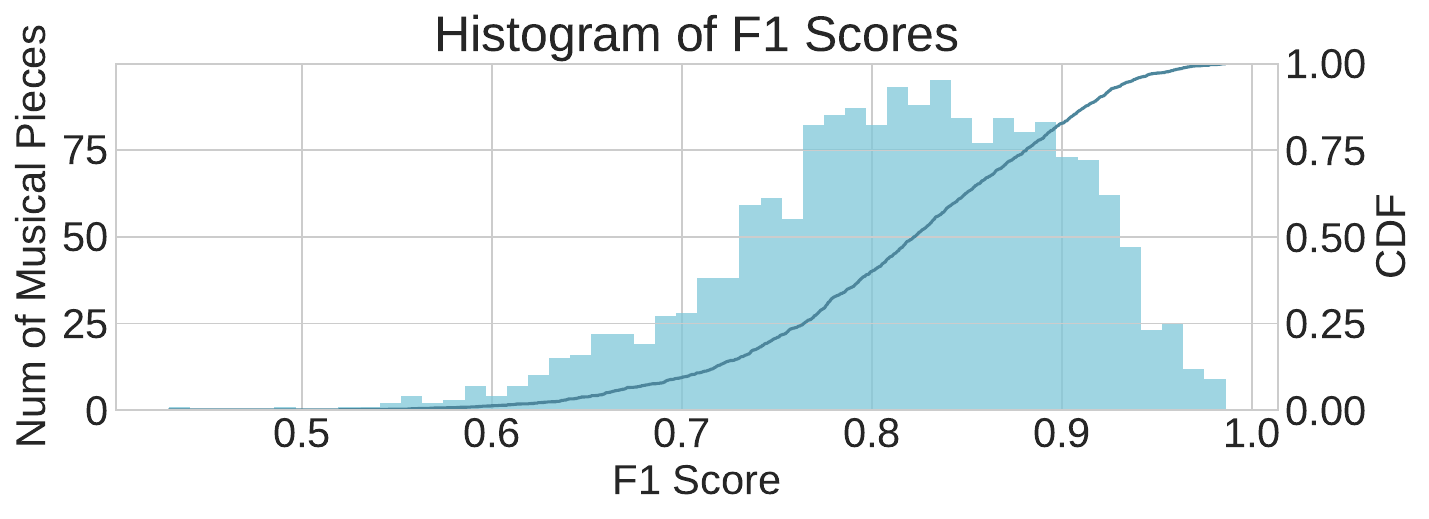}
\caption{Statistics of our RP1M / RP1M++ dataset. (\textbf{Top}) Histogram of pressed keys in our dataset. (\textbf{Bottom Left}) Distribution of the number of active keys over all time steps. (\textbf{Bottom Right}) Distribution of F1 scores of RL agents used to collect the dataset.}
\label{fig:dataset}
\vspace{-0.3cm}
\end{figure*}

In Fig.~\ref{fig:dataset}, we show the statistics of our collected motion dataset. The top plot shows the histogram of the pressed keys among all MIDI files used for RL specialists training. We found that keys close to the center are more frequently pressed than keys at the corners. Also, white keys, taking 65.7\%, are more likely to be pressed than black keys. In the bottom left plot, we show the distribution of the number of active keys over all time steps.
It roughly follows a Gaussian distribution, and 90.70\% of musical pieces in our dataset include 1000-4000 active keys. We also include the distribution of F1 scores of trained agents used for collecting data. We found that most agents (79.00\%) achieve F1 scores larger than 0.75, and 99.89\% of the agents' F1 scores are larger than 0.5. 
The distribution of F1 scores reflects the quality of the collected dataset. We empirically found agents with F1 scores $\geq$ 0.75 are capable of playing sheet music reasonably well. Agents with $\leq$ 0.5 F1 scores usually have notable errors due to the difficulty of songs or the mechanical limitations of the Shadow robot hand. We also include the F1 scores for each piece in our dataset so users can filter the dataset according to their needs. 

\subsection{RP1M++ Dataset}
Although the large-scale \textit{RP1M} dataset contains approximately one million expert trajectories across $\sim$2,000 musical pieces, its primary limitation lies in the lack of intra-song diversity.
Each piece provides 500 trajectories generated by rolling out the same trained RL policy.
While actions are stochastically sampled from the expert policy, we observe that after sufficient training, the policy distribution becomes narrow, yielding highly similar trajectories.
This limited variation reduces the effectiveness of imitation learning: once a behavior-cloning agent deviates from the narrow expert distribution, subsequent states often drift off the training manifold, causing severe performance degradation.

To address this issue, we construct a more diverse dataset, \textit{RP1M++}.
Like RP1M, it covers $\sim$2,000 music clips, but substantially increases intra-song diversity by labeling replay-buffer data using DAgger~\cite{ross2011reduction}.
Specifically, for each clip, we load a pre-trained expert policy $\pi_{\text{exp}}$ and randomly initialize a student policy $\pi_{\text{stu}}$.
The student interacts with the environment to collect trajectories $\tau_i(s_t, a_t)$, which are then relabeled by the expert $\pi_{\text{exp}}(\cdot|s_t)$ to produce $\tau_i(s_t, \hat{a}_t)$.
The student is trained with standard behavior cloning on these relabeled trajectories, and this process iterates until $\pi_{\text{stu}}$ matches $\pi_{\text{exp}}$.
Consequently, the relabeled dataset contains states visited by a wide range of intermediate policies obtained during the training of $\pi_{\text{stu}}$, resulting in a more diverse state distribution than using only the expert policy.
The resulting relabeled trajectories $\tau_i(s_t, \hat{a}_t)$ are saved to form RP1M++.
To accelerate training, we pre-fill the replay buffer with five expert trajectories to warm up $\pi_{\text{stu}}$.
A detailed description of the algorithm is given in~\cref{alg:data_collection}.
Compared to storing the full relabeled RL replay buffer, our dataset is much more compact.
Moreover, unlike directly applying DAgger for training the multi-song agent, our pipeline parallelizes easily and simplifies the training loop of the multi-task learner.

\begin{algorithm}[t]
\caption{Data Collection Pipeline of RP1M++}\label{alg:data_collection}
\begin{algorithmic}
\Require RL expert policy $\pi_{\text{RL}}$, student agent $\pi_{\phi}$, replay buffer $\mathcal{D}_{\text{}} \gets \emptyset$, target episodic return $R_{\text{tar}}$, current episodic return $R_{\text{cur}} \gets 0$.

\While{$R_{\text{cur}} \leq R_{\text{tar}}$}
\State \textcolor{vibrantblue}{\emph{// Collect Relabeled Trajectories}}
    \State $t\gets 1, ~R_{\text{cur}}=0$
    \While{$t \leq \text{episode length}$}
        \State Obtain action $a_t \sim \pi_{\phi}(a_t|o_t)$ with the student policy.
        \State Interact the env. with $a_t$ to obtain $(o_t, a_t,r_t,o_{t+1})$.
        \State Relabel action with $\tilde{a}_t = \pi_{\text{RL}}(a_t|o_t)$
        \State Store transition $\mathcal{D} \gets (o_t, \textcolor{vibrantred}{\tilde{a}_t},r_t,o_{t+1})$
        \State Calculate return $R_{\text{cur}} \gets R_{\text{cur}}+r_t$
    \EndWhile

    \State \textcolor{vibrantblue}{\emph{// Update Student Policy $\pi_\phi$}}
    \For{G grad steps}
        \State Randomly sample a mini-batch $\mathcal{B}= \{ o_i, a_i\}_{i=0}^{|\mathcal{B}|}$
        \State Update $\pi_\phi$ by minimizing $-\frac{1}{|\mathcal{B}|}\sum_{i=0}^{|\mathcal{B}|} \log \pi_\phi(a_i|o_i)$.
    \EndFor
\EndWhile

\end{algorithmic}
\end{algorithm}

\section{Training Multi-Song Piano-Playing Agents}
This section introduces \textsc{OmniPianist}, our multi-song piano-playing agent.
\textsc{OmniPianist} is built on a Flow Matching Transformer, trained on the RP1M++ dataset (see~\cref{sec:dataset}) using scalable imitation learning.
We begin by introducing flow matching in~\cref{sec:fm}, with a focus on the construction of the training objective used for \textsc{OmniPianist}.
We then describe the training procedure of \textsc{OmniPianist} in~\cref{sec:fm_training}.

\subsection{Flow Matching Policy} \label{sec:fm}

\begin{figure*}[t] 
\centering
 \setlength{\abovecaptionskip}{0.1cm}
 \includegraphics[width=0.98\textwidth]{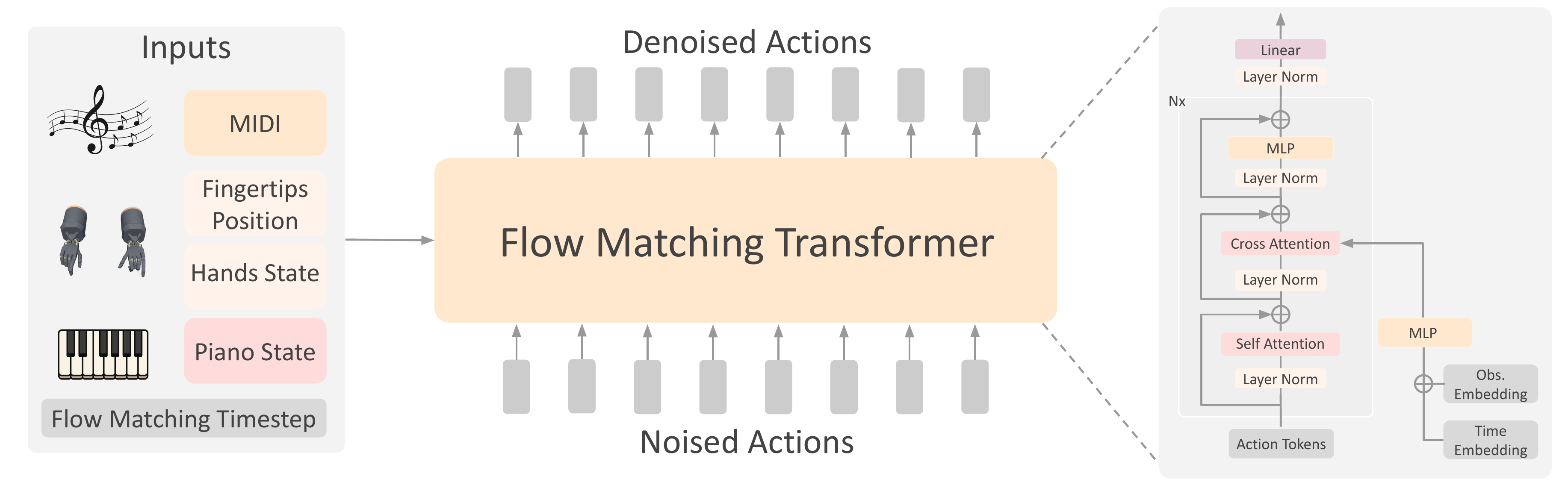}
 
\caption{Flow Matching Transformer architecture. The transformer is conditioned on the goal extracted from MIDI files, robot hand fingertip positions, robot hand states, and piano states. Noised action tokens are linearly embedded, combined with learned positional embeddings, and passed through an $N$-block stack of Transformer decoder layers ($N=12$ in our case). The flow matching timestep is encoded using a sinusoidal embedding and concatenated with linearly projected observation tokens to form the conditioning sequence. This sequence is processed by a lightweight per-token MLP and serves as the cross-attention memory for the decoder. Finally, a layer norm and linear head map the token features to continuous actions. }
\label{fig:architecture}
\end{figure*}

Flow matching~\cite{lipman2022flow,pooladian2023multisample} is a generative modeling framework that combines aspects from Continuous Normalising Flows (CNFs) and diffusion models described in~\cref{sec:pre}, offering a more direct and stable alternative. It learns to transform a simple prior distribution (e.g., Gaussian noise) into a complex target distribution by directly matching the vector fields $v_t$ to map samples from the source distribution $p$ to the target data distribution $q$.
Unlike diffusion models, which gradually add and remove noise through discrete steps, flow matching defines a continuous-time transport process governed by a learned Ordinary Differential Equation (ODE), enabling exact likelihood computation and efficient sampling. Compared to diffusion models, flow matching is simpler, using fewer hyperparameters, and offers greater numerical stability, having outperformed them across a range of domains, making it a compelling alternative.

Flow matching constructs a target vector field that defines a flow by solving an ODE. The flow deterministically connects points from the prior distribution to corresponding data samples. A neural network, parameterized by $\theta$, is then trained to predict this vector field. More specifically, the goal is to find the parameters of the flow defined as a learnable velocity $u_\theta$ that generates intermediate distributions $p_t$ with $p_0 = p$ and $p_1 = q$ for each $t \in [0, 1]$. The flow matching loss measures how well the learned velocity field aligns with the ideal velocity field along the probability path from the source to the target distribution. The training objective is typically a straightforward regression loss:
\begin{equation}
    \mathcal{L}(\theta) = \mathbb{E}_{t, x \sim p_t} \left[ \left\| u^{\theta}_t(x, t) - v_t(x) \right\|^2 \right]
\end{equation}
where $x$ represent a sample from the intermediate distribution $p_t$ at time $t\sim \mathcal{U}[0, 1]$, $v_t(x)$ is the target vector at that point, and $u^{\theta}_{t}(x,t)$ is the network's prediction. This loss function encourages the learned velocity field $u^{\theta}_{t}$ to match the ideal velocity field $v_t$, effectively guiding samples along the desired probability path.
However, computing the exact velocity field $v_t$ for the entire dataset is often computationally infeasible. To address this, \cite{lipman2022flow} introduced \textit{Conditional Flow Matching} (CFM), which reformulates the problem by conditioning on individual data points, making it possible to efficiently train a parametric vector field $u^{\theta}$ using the following CFM loss:
\begin{equation}
    \mathcal{L}(\theta) = \mathbb{E}_{t,\, x_1 \sim q,\, x_t \sim p_t(x|x_1)} 
    \Big[ \, \big\| u^{\theta}_t(x_t, t) - v_t(x_t \mid x_1) \big\|^2 \, \Big],
\end{equation}
where $x_1 \sim q$ is the conditioning variable (e.g., a specific data sample or observation), and $x_t$ is sampled from the conditional distribution $p_t(\cdot \mid x_1)$. The ideal conditional velocity field $v_t(x \mid x_1)$ can be simplified as $x_1 - x_0$, where $x_0 \sim p_0$ is drawn from a simple base distribution and $x_1 \sim p_1$ from the target distribution.  
Following Optimal Transport theory, for a randomly chosen $t \in [0,1]$, an interpolated point is defined as $x_t = (1-t)x_0 + t x_1$.
The model is then trained to predict the velocity that moves $x_t$ in the correct direction by minimizing:
\begin{equation}
    \mathcal{L}(\theta) = \mathbb{E}_{t,~ x_0\sim p_0,~ x_1\sim p_1} 
    \Big[ \, \big\| u^{\theta}_t(x_t, t) - (x_1-x_0))) \big\|^2 \, \Big],
\end{equation}

Similar to Diffusion Policy, extending the principles of flow matching to imitation learning gives rise to the \textit{Flow Matching Policy}~\cite{black2410pi0,zhang2024affordance}. This approach frames the robot policy $\pi_\theta(a|s)$ as a conditional normalizing flow (CNF) trained to generate action sequences $a$ conditioned on observations $s$. The policy learns a conditional vector field $u_\theta(a, t |s)$ that continuously transports an initial noise sample $a_0 \sim \mathcal{N}(0, \mathbf{I})$ to an expert action $a_1$ over a time horizon $t \in [0, 1]$.  

During training, Flow Matching Policy employs a behavior cloning objective based on the flow matching loss:
\begin{equation}
    \mathcal{L}(\theta) = \mathbb{E}_{t,~ a_0\sim p_0,~ a_1\sim p_1} 
    \Big[ \, \big\| u^{\theta}_t(a_t, t | s) - (a_1-a_0))) \big\|^2 \, \Big],
    \label{eq:obj}
\end{equation}
where the model learns to predict the velocity vector that moves an interpolated action $a_t$ toward the ground-truth expert action $a_1$.
At inference time, the policy generates actions by initializing from random noise and integrating the learned ODE using a numerical solver:
\begin{equation}
    a_1 = a_0 + \int_0^1 u_\theta(a_t, t, s) \, dt.
\end{equation}

\subsection{Training \textsc{OmniPianist} with Flow Matching} \label{sec:fm_training}

We adopt the Flow Matching Transformer as the policy representation in OminiPianist, which is shown in Fig.~\ref{fig:architecture}. In our robotic piano-playing setting, the target trajectory $a_1$ in Eq.~\ref{eq:obj} denotes expert demonstration actions from the RP1M++ dataset, comprising joint positions of the robot hands, forearm positions, and sustain-pedal control. The source trajectory $a_0$ consists of randomly generated waypoints drawn from a standard multivariate normal, $a_0 \sim \mathcal{N}(\mathbf{0}, \mathbf{I})$. The policy is conditioned on the observation $s$, which includes the goal extracted from the MIDI file, the robot hand's fingertip positions, the robot hand's states, and the piano state. To provide short-horizon lookahead, the goal vector concatenates the current step goal with multiple future steps' goals.
Noised action tokens are linearly embedded, added with learned positional embeddings, and passed through a non-causal Transformer decoder. The flow matching timestep is encoded by a sinusoidal embedding and concatenated with linearly projected observation tokens to form a conditioning sequence; this sequence is processed by a small per-token MLP and serves as the keys/values that the decoder cross-attends to. Finally, a layer norm and linear head project token features to continuous actions. Unlike other parts that use BF16 for efficiency, attention computations use FP32 for stability.

\begin{figure*}[t] 
\centering
 \setlength{\abovecaptionskip}{0.1cm}
    \includegraphics[width=0.98\textwidth]{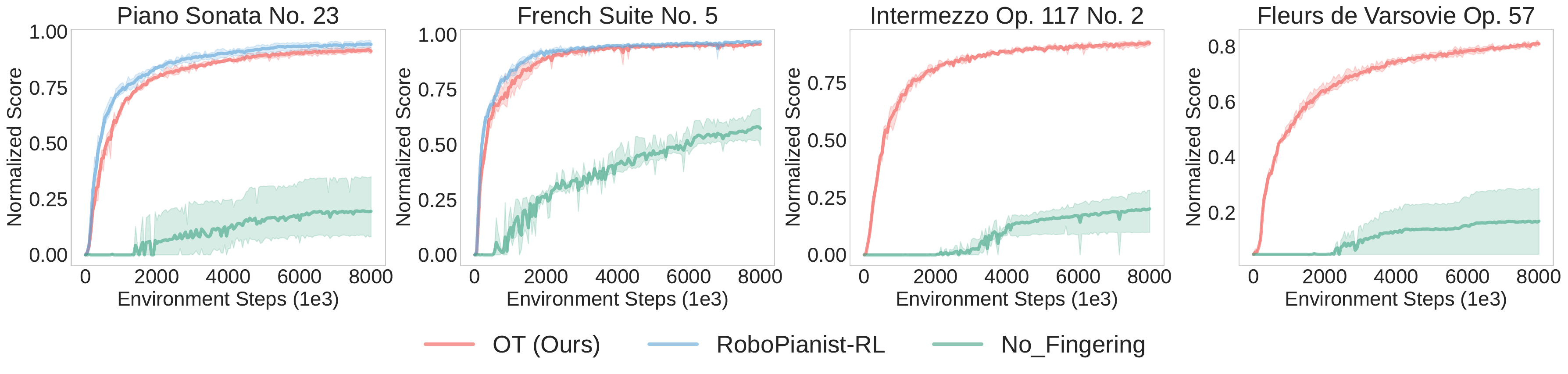}
\caption{
    Comparison of the RL performance with our OT fingering, human-annotated fingering, and no fingering.
    Our method matches the performance of RoboPianist-RL, which is trained with human fingering.
    Our method also outperforms the baseline without any fingering information by a large margin.
    The plots show the mean over 3 random seeds, and the shaded areas represent the 95\% confidence interval.
}
\label{fig:specialist_comparison_to_baselines}
\vspace{-0.3cm}
\end{figure*}

\begin{figure*}[t] 
\centering
 \setlength{\abovecaptionskip}{0.1cm}
\includegraphics[width=0.99\textwidth]{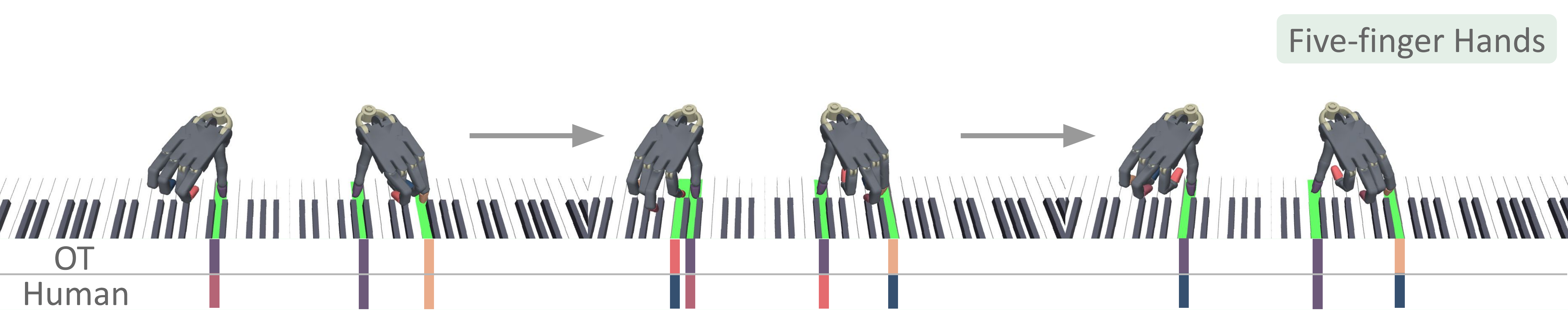}
\includegraphics[width=0.99\textwidth]{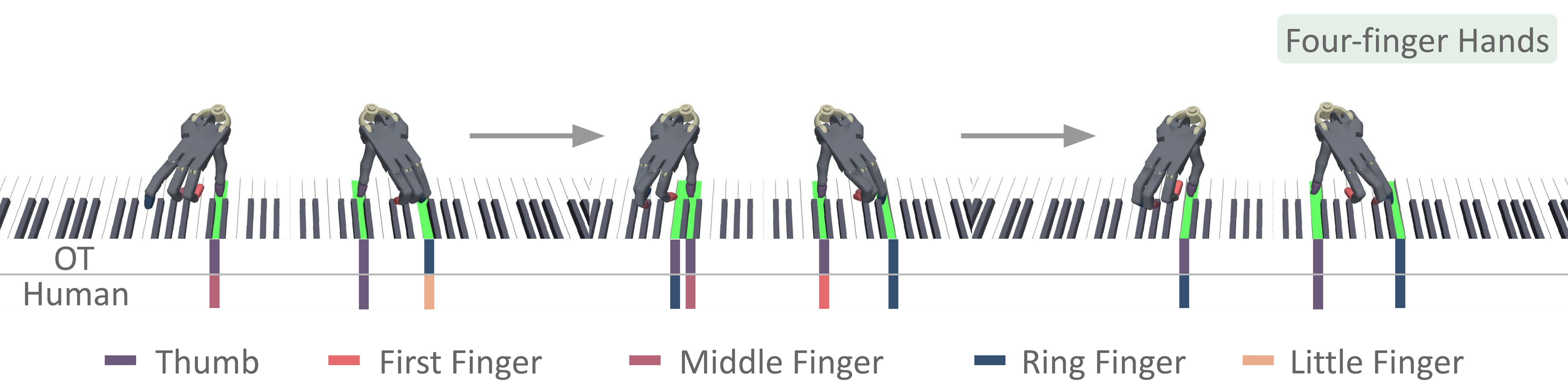}
\caption{Comparison of fingering discovered by the agent itself and human annotations. We visualize a sample trajectory of playing the French Suite No.5 Sarabande, along with the corresponding fingering. The agent discovers a fingering strategy that differs from human annotations, adapting to hardware constraints while accurately pressing the target keys.}
\label{fig:fingering}
% \vspace{-0.2cm}
\end{figure*}

\section{Single-song Specialist Evaluation}
In this section, we evaluate the piano-playing performance of the RL specialists introduced in \cref{sec:specialist}, and analyze both the learned fingering strategies and the agents’ performance across different robotic hands.

\subsection{Piano Playing Performance}

The performance of the specialist RL agents decides the quality of our dataset. In this section, we investigate the performance of our specialist RL agents. We are interested in i) how the proposed OT-based finger placement helps learning, ii) how the fingering discovered by the agent itself compares to human fingering labels, and iii) how our method transfers to other embodiments.

In Fig.~\ref{fig:specialist_comparison_to_baselines}, we compare our method with RoboPianist-RL both with and without human fingering. We use the same DroQ algorithm with the same hyperparameters for all experiments. 
RoboPianist-RL includes human fingering in its inputs, and the fingering information is also used in the reward function to force the agent to follow this fingering. Our method, marked as \emph{OT}, removes the fingering from the observation space and uses OT-based finger placement to guide the agent to discover its own fingering. We also include a baseline, called \emph{No Fingering}, that removes the fingering entirely.
The first two columns of Fig.~\ref{fig:specialist_comparison_to_baselines} show that our method without human-annotated fingering matches RoboPianist-RL's performance on two different songs. 
Our method outperforms the baseline without human fingering by a large margin, showing that the proposed OT-based finger placement boosts the agent's learning.
The proposed method works well even on challenging songs. We test our method on \emph{Flight of the Bumblebee} and achieve a 0.79 F1 score after 3M training steps. To the best of our knowledge, we are the first to play the challenging song Flight of the Bumblebee with general-purpose bimanual dexterous robot hands.

\subsection{Analysis of the Learned Fingering}

We now compare the fingering discovered by the agent itself and the human annotations. In~\cref{fig:fingering}, we visualize the sample trajectory of playing \emph{French Suite No.5 Sarabande} and the corresponding fingering. We found that although the agent achieves strong performance for this song (the second plot in~\cref{fig:specialist_comparison_to_baselines}), our agent discovers different fingering compared to humans. For example, for the right hand, humans mainly use the middle and ring fingers, while our agent uses the thumb and first finger. Furthermore, in some cases, human annotations are not suitable for the robot hand due to different morphologies. For example, in the second time step of~\cref{fig:fingering}, the human uses the first finger and ring finger. However, due to the mechanical limitation of the robot hand, it can not press keys that far apart with these two fingers; thus, mimicking human fingering will miss one key. Instead, our agent discovered to use the thumb and little finger, which satisfies the hardware limitation and accurately presses the target keys.

\subsection{Cross Embodiments}

\begin{figure}[t]
\centering
 \setlength{\abovecaptionskip}{0.1cm}
 \includegraphics[width=0.48\textwidth]{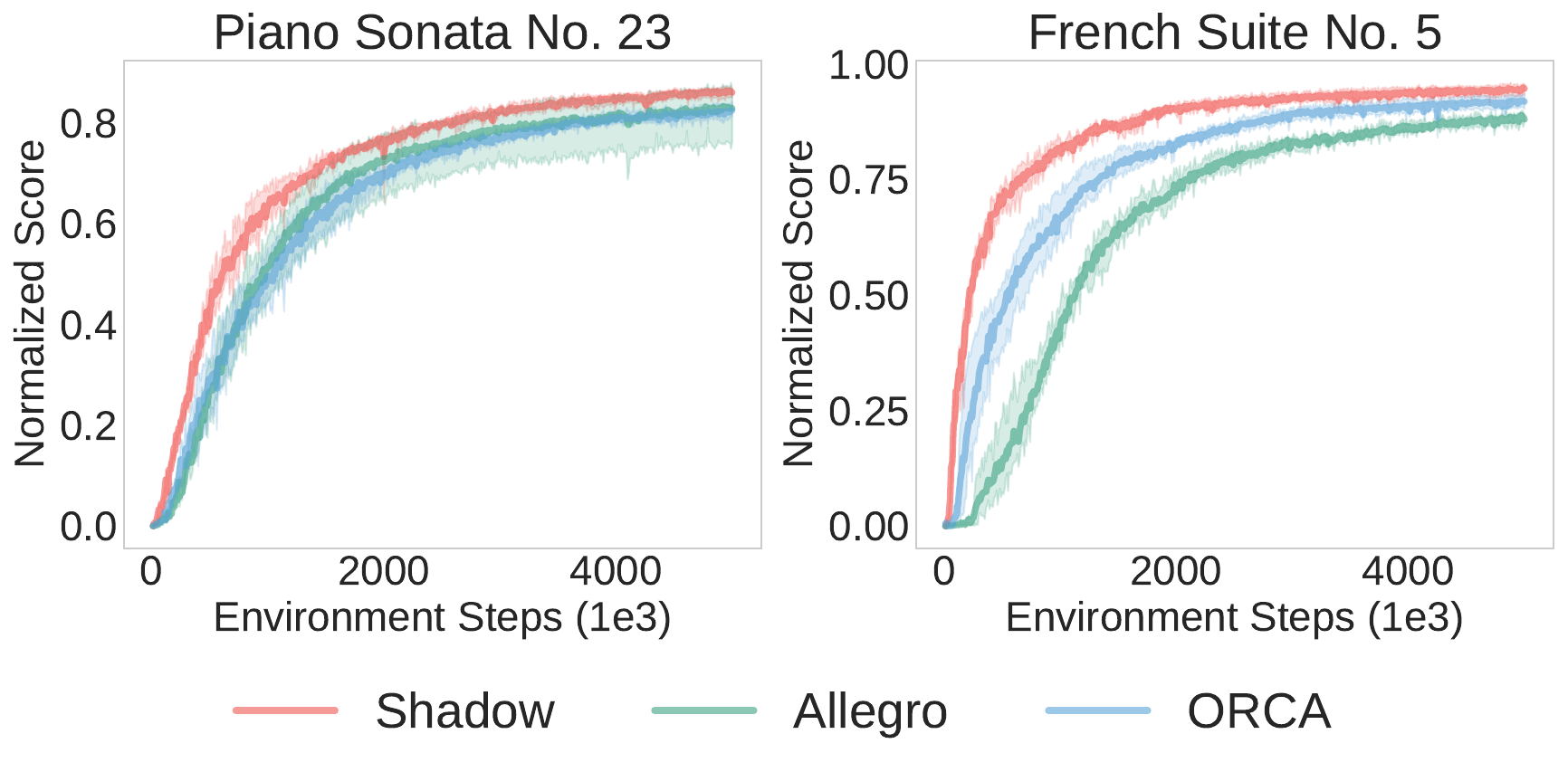}
\includegraphics[width=0.48\textwidth]{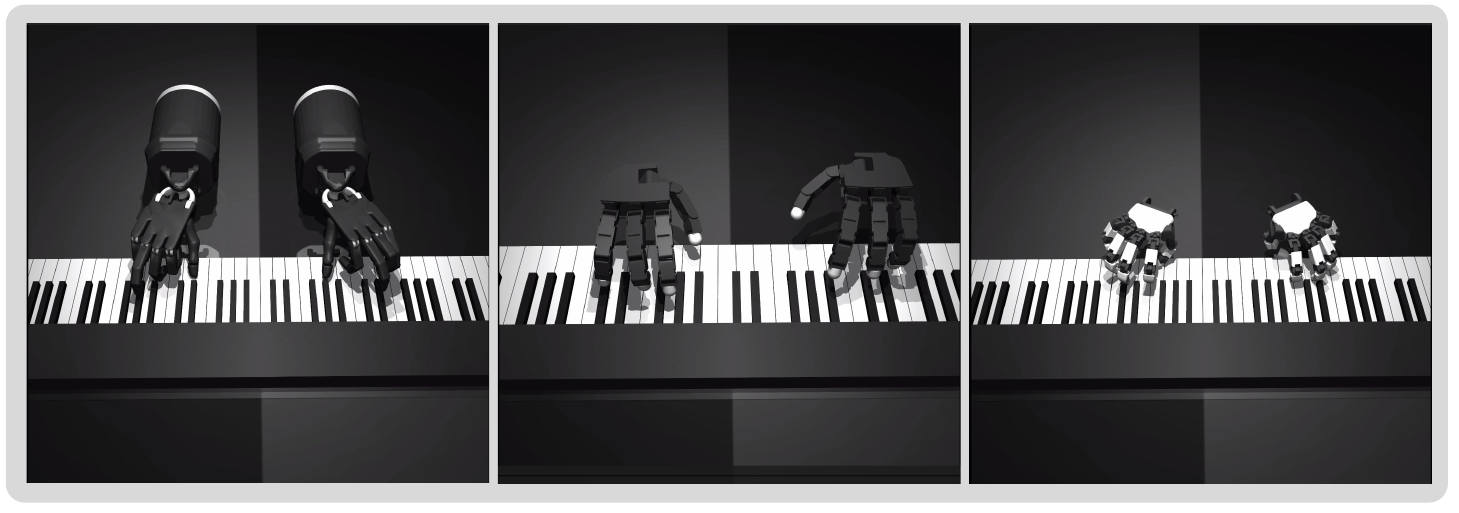}
\caption{Cross-Embodiment experiments. We have investigated three different robot hands in simulation for piano playing—from left to right: Shadow, Allegro, and Orca Hands. As illustrated by the performance curves, our automatic fingering method enables the successful training of piano-playing policies using RL across all three embodiments.}
\label{fig:x_embodiment}
\vspace{-0.4cm}
\end{figure}

Research laboratories usually have different robot platforms; thus, having a method that works for different embodiments is highly desirable. We test our method on a different embodiment.
We apply our method on Allegro~\cite{allegro} and ORCA~\cite{christoph2025orca} hands.
We evaluate the different robot hands on the Piano Sonata No.23 as well as the French Suite No.5.
As shown in~\cref{fig:x_embodiment}, although different hands have different learning progress, all hands can converge to similar performance.
Due to the mechanical constraints, the agent discovers different fingering compared to humans and the Shadow hands, but still accurately presses active keys, meaning our method is compatible with different embodiments.

\section{Multi-Song Learner Evaluation}
 
In this section, we evaluate the performance of our multi-song learning agent, \textsc{OmniPianist}. To be specific, the objective is to train a single multi-task policy capable of playing various music pieces on the piano. 
We train the policy on a portion of the RP1M++ dataset and evaluate its in-distribution performance (F1 scores on songs included in the training data) and its generalization ability (F1 scores on songs not present in the training data).
We evaluate and compare policies across three dimensions: the performance of different policy representations, the impact of training on distinct datasets, and the effect of scaling to larger amounts of training data.

\textbf{Training and Evaluation Setup}~~Both the RP1M and RP1M++ datasets contain 1,053 unique songs and 2,091 music clips, with some songs segmented into shorter chunks to maintain a consistent episode length. We train policies using expert trajectories from progressively larger subsets of the RP1M++ dataset, consisting of 12, 150, 300, 500, 700, and 900 songs. Each smaller subset is a strict subset of the larger ones, enabling controlled analysis of data scale. To evaluate policy performance, we use two distinct sets of musical pieces. The first set consists of \textbf{in-distribution} songs—those included in the training data. This evaluation assesses the policy’s multi-task learning ability and its capacity to accurately recall and reproduce songs it was trained on. Specifically, we use the RoboPianist-ETUDE-12 songs from the PIG dataset~\cite{nakamura2020statistical}. These songs are also included in the 12-song training subset. The second set of songs for evaluation contains \textbf{out-of-distribution} songs, music pieces that do not overlap with the training songs. These pieces are chosen for their diversity in motion and longer temporal horizons, making them particularly challenging to perform and suitable for evaluation. This out-of-distribution evaluation measures the zero-shot generalization capabilities of the policies. Analogous to an experienced human pianist who can play new pieces at first sight, we aim to determine if it is feasible to develop a piano-playing agent that can perform novel pieces under diverse and unseen conditions. We use 100 songs from the GiantMIDI-Piano dataset~\cite{kong2020giantmidi} for out-of-distribution evaluation.

\textbf{Policy Representations}~~Piano playing is a challenging task that combines various aspects of dynamic and manipulation functions. To capture the complexity of our large-scale motion dataset, we need an expressive policy representation. We evaluated Diffusion Policy~\cite{chi2023diffusion} with U-Net~\cite{ronneberger2015u} (DDIM), Flow Matching Policy with U-Net~\cite{zhang2024affordance} (FM), and Flow Matching Policy with Transformer~\cite{vaswani2017attention} (FMT). The Diffusion Policy we used is based on the DDIM~\cite{song2020denoising} scheduler, and the Flow Matching Policy is based on the Euler discrete scheduler~\cite{esser2024scaling}. 
The U-Net-based policies incorporate Feature-wise Linear Modulation (FiLM)\cite{perez2018film} for encoding observations.

\subsection{Datasets Comparison}

\begin{figure*}[t]
\centering
\includegraphics[width=1.0\textwidth]{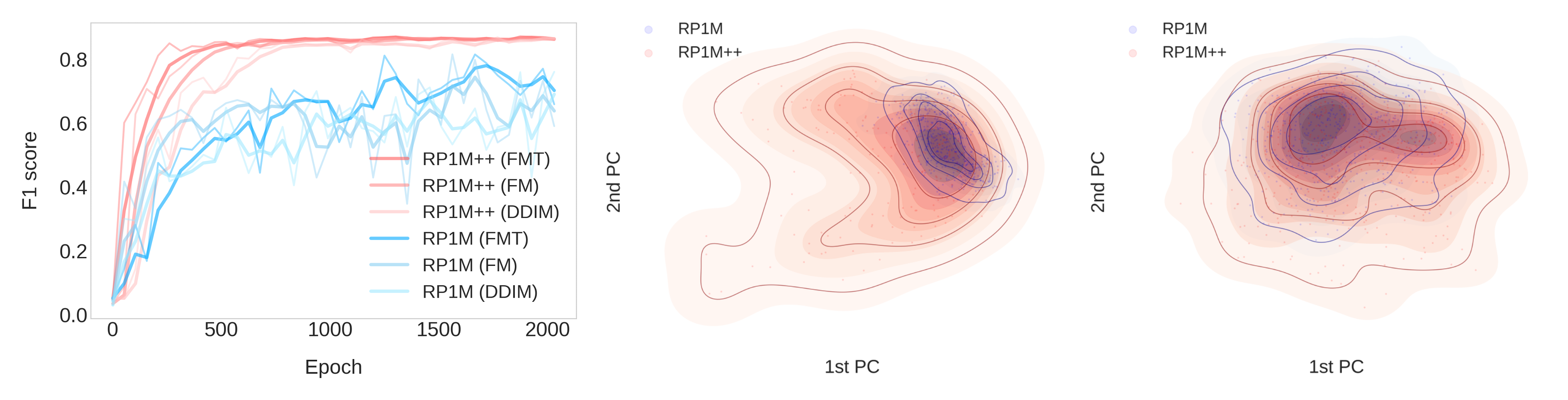}
{\setlength{\abovecaptionskip}{0.1cm}%
\caption{Comparison of RP1M and RP1M++. (\textbf{Left}) F1-score curves when training different policies on each dataset; policies trained on RP1M++ consistently outperform those trained on RP1M. (\textbf{Middle}) KDE plot for RP1M (blue) and RP1M++ (orange), showing the trajectory distributions for \emph{Golliwogg's Cakewalk}. Each point is a trajectory's average hand-joint observations over time, projected onto the first two Principal Components (PC) obtained on the combined set (axes: 1st PC and 2nd PC). Light scatter shows individual trajectories, while contours depict density. (\textbf{Right}) KDE plot for \emph{French Suite No.\,5} (Sarabande). RP1M++ spans a broader and more diverse region of the state space.}
\label{fig:dataset_comparison}
}
\vspace{-0.3cm}
\end{figure*}

In this section, we compare the effectiveness of the RP1M++ dataset described in~\cref{sec:dataset} with the original RP1M dataset~\cite{zhao2024rp1m}. To ensure a fair comparison, we use the same 12-song subset from RP1M++ and select the corresponding songs from RP1M. We then train separate policies on each dataset, and evaluate their in-distribution performance, allowing us to evaluate the impact of the enhanced data quality and diversity in RP1M++.

\textbf{Results}~~
As show in the left figure of~\cref{fig:dataset_comparison}, we compare the F1 scores of three different policies, including FMT, FM and DDIM, trained on both RP1M, and RP1M++. 
% The results are presented in Fig.~\ref{fig:dataset_comparison}. The plot on the left shows the F1 scores of all policies trained on each dataset. 
From these curves, we can see that policies trained on RP1M++ consistently outperform those trained on RP1M. In addition to improved performance, training on RP1M++ also exhibits greater stability across runs. This improvement is largely attributed to the increased diversity in the state distribution provided by the RP1M++ dataset. This is further illustrated in the Kernel Density Estimation (KDE) plots shown in the middle and right panels of Fig.~\ref{fig:dataset_comparison}, where each point represents a trajectory. As shown in the KDE plot, we visualize each trajectory after projecting the per-trajectory state statistics onto the first two principal components (1st PC1, 2nd PC). The orange RP1M++ density exhibits broader support
compared to the blue RP1M density, which is more compact and centrally concentrated. While the two datasets overlap in the high-density core, RP1M++ spans a noticeably broader and more diverse region of the state space, indicating greater variability and wider coverage in the data. It reduces distributional blind spots during training, leading to more stable optimization and better generalization on out-of-distribution songs.

\subsection{Policy Representations Comparison}

To compare the effectiveness of different policy representations, we train each model on two training set sizes, 12 songs and 300 songs, and evaluate their F1 scores on both in-distribution and out-of-distribution songs, as previously described. 

\begin{figure}[] 
\centering
 \setlength{\abovecaptionskip}{0.1cm}
 \includegraphics[width=0.49\textwidth]{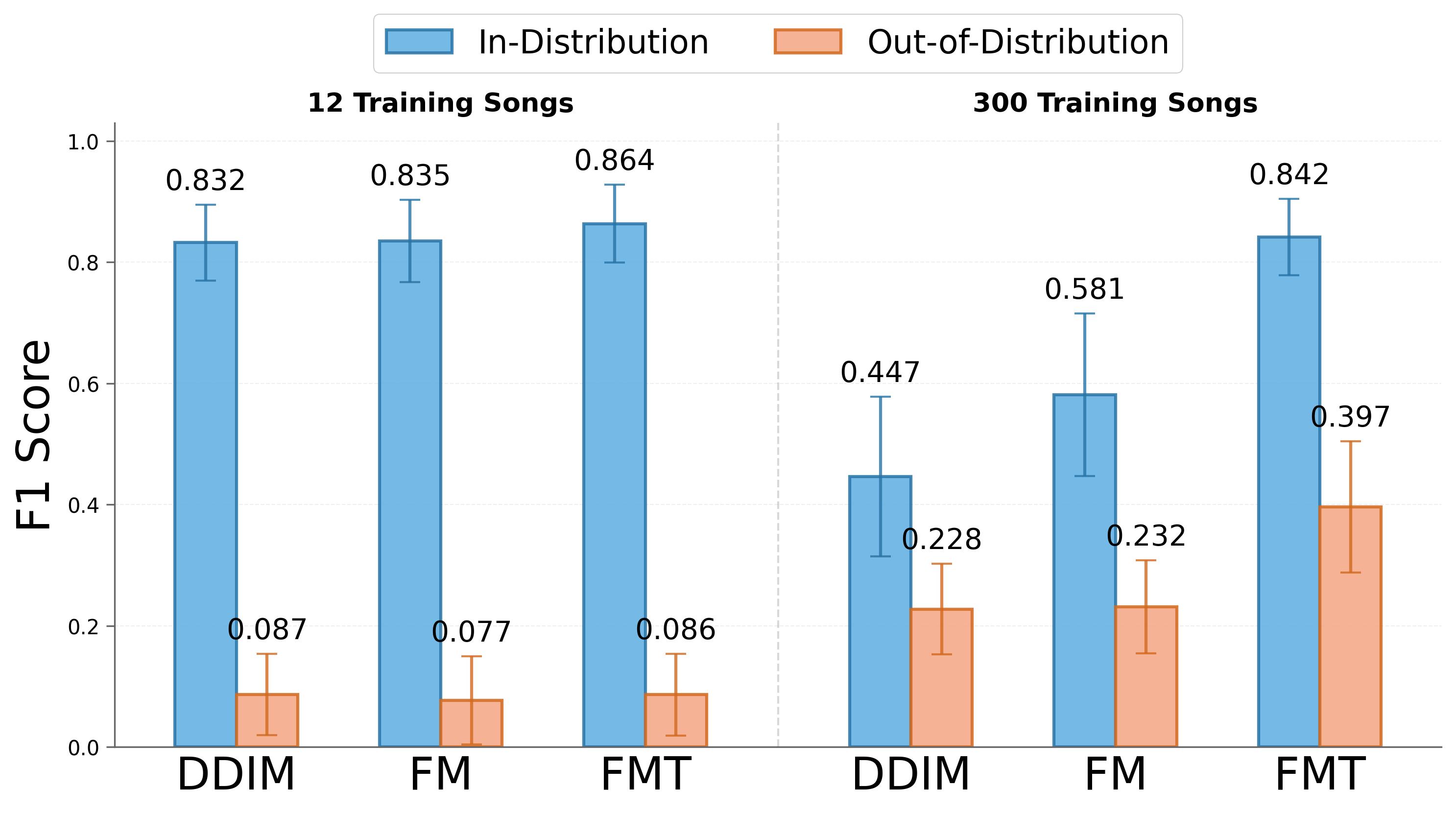}
\caption{Comparison of policy representations on in‑distribution (blue) and out‑of‑distribution (orange) evaluation. Bars show mean F1 as numeric labels with task‑level standard deviation as error bars. Policies (DDIM, FM, FMT) are grouped by training set size (12 vs 300 songs). Higher values indicate better performance.}
\label{fig:policy_representation}
\vspace{-0.2cm}
\end{figure}

\textbf{Results}~~
% The results are summarized in Fig.~\ref{fig:policy_representation}. 
In~\cref{fig:policy_representation}, we compare the performance of different policy representations on both in-distribution and out-of-distribution songs, using models trained separately on the 12-song and 300-song subsets.
When trained on the 12-song subset, all policy representations achieve high F1 scores on the in-distribution songs, indicating their ability to fit the training data effectively. However, their performance on out-of-distribution songs remains low, which is expected given the limited diversity and coverage of the 12-song training set. Clear differences between policy representations become obvious when training on the more diverse 300-song dataset. Among those policy representations, the Flow Matching Transformer (FMT) achieves the highest F1 scores on both in-distribution and out-of-distribution evaluations, significantly outperforming the UNet-based Flow Matching Policy (FM) and the Diffusion Policy (DDIM). These results highlight FMT’s superior expressiveness and generalization capability. While the UNet-based Flow Matching Policy shows better performance than the Diffusion Policy, both of them still struggle to fully capture the complexity of the larger dataset—evidenced by a drop in its in-distribution F1 score compared to training on only 12 songs. In contrast, the Flow Matching Transformer not only maintains strong in-distribution performance but also exhibits a substantial improvement in out-of-distribution generalization when scaled up to 300 training songs. This suggests that FMT can better handle both the richness and variability present in large-scale piano-playing data.

\subsection{Data Scaling Law}

\begin{figure}[] 
\centering
 \setlength{\abovecaptionskip}{0.1cm}
 \includegraphics[width=0.49\textwidth]{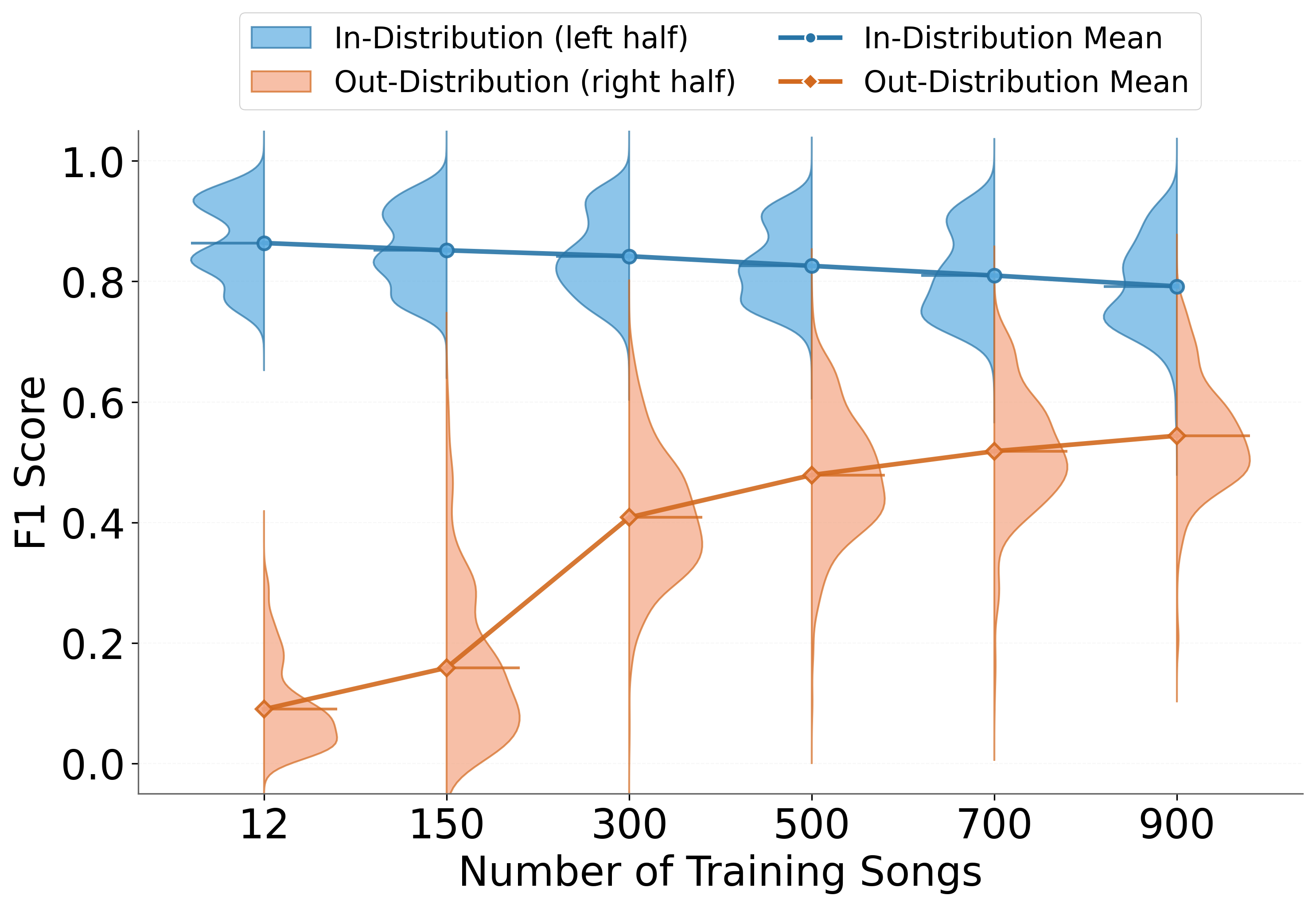}
\caption{Data scaling law. For each training‑song count on the x‑axis, split violins summarize the distribution of per‑task F1 scores: in‑distribution on the left half (blue) and out‑of‑distribution on the right half (orange). Overlaid solid lines with circle (in‑distribution) and diamond (out‑of‑distribution) markers connect the mean F1 across training sizes. The y‑axis reports F1 scores; higher is better.}
\label{fig:scale_law}
\vspace{-0.2cm}
\end{figure}

The policy representation comparison indicates that the Flow Matching Transformer substantially outperforms the alternatives, suggesting greater expressiveness and a stronger capacity to model the complexity of RP1M++. In this section, we are going to conduct the data scaling law experiment. As described earlier, we train the Flow Matching Transformer policies on progressively larger subsets of RP1M++ containing 12, 150, 300, 500, 700, and 900 songs, where each smaller subset is a strict subset of the larger ones to enable controlled analysis. Our goal is to quantify how performance scales with data by tracking both in-distribution and out-of-distribution F1 scores as training data increases.

\textbf{Results}~~
% The results are shown in Fig.~\ref{fig:scale_law}.
\Cref{fig:scale_law} shows the split violin plot summarizes F1 scores for in-distribution and out-of-distribution evaluations as we scale the number of training songs from 12 to 900. For in-distribution evaluation, the tight blue violins and small error bars indicate low variance across seeds and songs, suggesting that the policy reliably memorizes or reconstructs training pieces even as the task set broadens. The in-distribution performance remains high and stable across all settings, with only a slight downward drift in the mean from $\sim$0.86 (12 songs) to $\sim$0.80 (900 songs), suggesting a small trade-off between memorization of the training pieces and broader competence as the task set grows. In contrast, the orange violins shift upward steadily as data increases, showing that out-of-distribution performance improves monotonically with data. The progressively narrowing orange violins and shrinking error bars further indicate increased stability and reliability in zero-shot generalization as we add data. Scaling data improves both typical and worst-case generalization. Together, these trends show that adding diverse songs substantially boosts out-of-distribution accuracy and stabilizes results while maintaining strong in-distribution performance.

\section{Conclusion}
In this paper, we presented \textsc{OmniPianist}, a Flow Matching Transformer capable of performing nearly one thousand music pieces. \textsc{OmniPianist} is trained on RP1M++, a large-scale motion dataset collected using more than 2,000 RL specialist agents. To enable scalable RL training at low cost, we removed the need for human-annotated fingering by introducing an automatic finger placement strategy based on Optimal Transport. Our approach is scalable, demonstrates strong generalization, and marks a step toward robotic agents with human-level dexterity in complex musical tasks.

Looking ahead, several directions remain open. First, our current evaluation based on F1 score does not fully capture the nuances of musical performance, as it only measures correct key presses; incorporating force and velocity could provide a more faithful assessment. Second, the RP1M++ dataset relies solely on proprioceptive observations, whereas humans leverage multimodal inputs such as vision, touch, and hearing; integrating these modalities may further enhance capability. Finally, deploying \textsc{OmniPianist} on real robots remains challenging, requiring accurate state estimation of both piano and hands, precise high-speed position control, and bridging the sim-to-real gap in highly dynamic tasks. 
Given the precise and dynamic nature of robotic piano playing, we believe this work lays a solid foundation for advancing contact‑rich, dynamic manipulation and for the development of dexterous robotic hands.

\section*{Acknowledgments}
We thank the Max Planck Institute for Intelligent Systems, Tübingen (Germany), for the support. 
We acknowledge CSC – IT Center for Science, Finland, for awarding this project access to the LUMI supercomputer, owned by the EuroHPC Joint Undertaking, hosted by CSC (Finland) and the LUMI consortium through CSC. 
Yi Zhao, Juho Kannala, and Joni Pajarinen acknowledge funding by the Research Council of Finland (352788, 362407, 345521, 353138, 357301). 
We thank Yuxin Hou and Wenyan Yang for the insightful discussion.


% \newpage
\bibliographystyle{IEEEtran}
\bibliography{references}

% \newpage

 

% \end{IEEEbiography}

\vspace{11pt}


\begin{IEEEbiography}[{\includegraphics[width=1in,height=1.25in,clip,keepaspectratio]{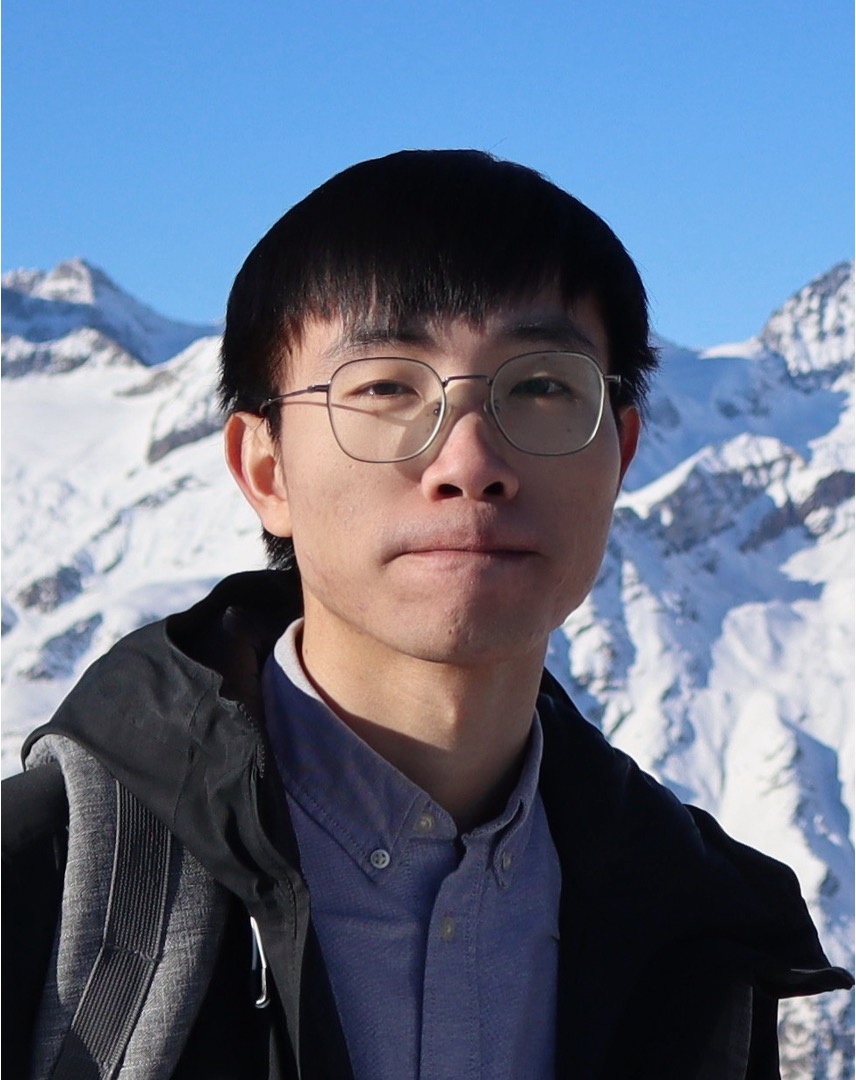}}]{Le Chen}
received the M.Sc. degree in Electrical Engineering and Information Technology from ETH Zurich, Switzerland, in 2023. He is currently pursuing a Ph.D. degree at the Empirical Inference Department of the Max Planck Institute for Intelligent Systems in Tübingen, Germany, as part of the ELLIS Ph.D. program. His research interests are in robotics and machine learning, with a long-term goal to build generalist agents and robots that can interact safely and reliably with humans and the physical world.
\end{IEEEbiography}
\vspace{-0.6cm}

\begin{IEEEbiography}[{\includegraphics[width=1in,height=1.25in,clip,keepaspectratio]{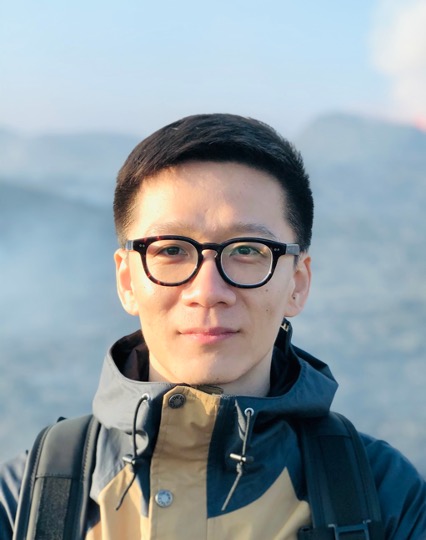}}]{Yi Zhao}
received his M.Sc. degree in Electrical Engineering and Automation from Aalto University, Finland, in 2020. He is currently a Ph.D. candidate in the Robot Learning Group at Aalto University, supervised by Joni Pajarinen and Juho Kannala. 
In 2024, he was a visiting researcher at the Empirical Inference Department of the Max Planck Institute for Intelligent Systems in Tübingen, Germany.
His research interests include reinforcement learning and imitation learning, with a focus on enabling robots to learn versatile skills in a sample-efficient manner.
\end{IEEEbiography}
\vspace{-0.6cm}

\begin{IEEEbiography}[{\includegraphics[width=1in,height=1.25in,clip,keepaspectratio]{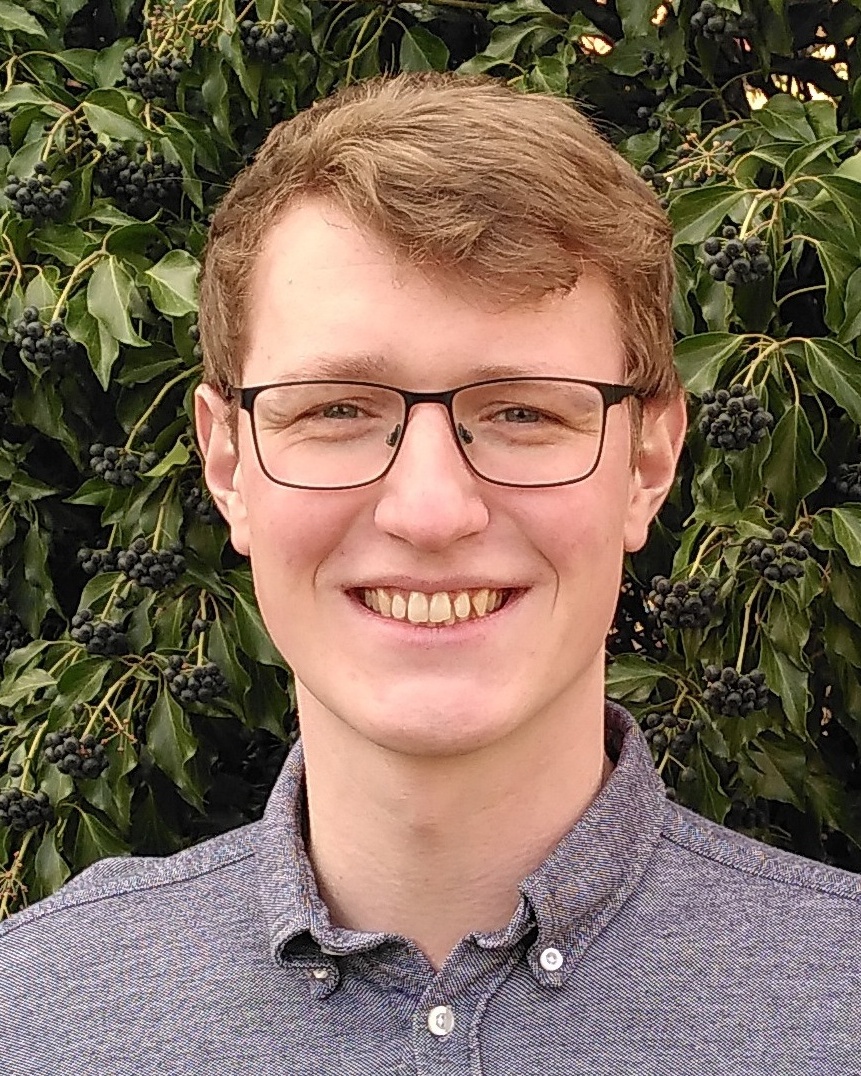}}]{Jan Schneider}
received the B.Sc. and M.Sc. degrees in computer science from Technical University of Darmstadt, Germany, in 2018 and 2022, respectively. He is currently pursuing a Ph.D. degree at the Empirical Inference Department of the Max Planck Institute for Intelligent Systems in Tübingen, Germany, as part of the ELLIS Ph.D. program.
His research interests lie in reinforcement learning for robotics, with a focus on action representations, learning dynamics, and sim-to-real transfer.
\end{IEEEbiography}
\vspace{-0.6cm}

\begin{IEEEbiography}[{\includegraphics[width=1in,height=1.25in,clip,keepaspectratio]{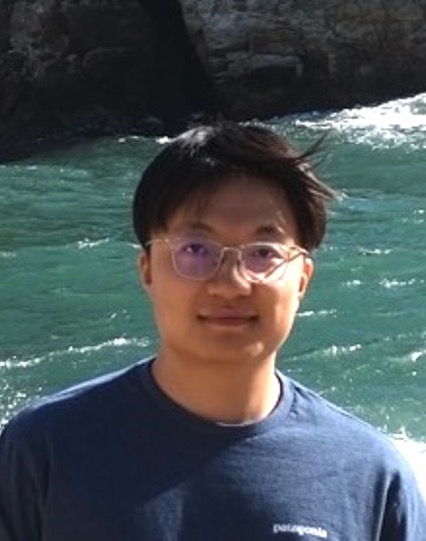}}]{Quankai Gao}
received the M.Sc. in Computer Science from University of Southern California and the B.E. in Automation Science and Engineering from South China University of Technology. He is now pursuing a Ph.D. degree in the Computer Science department at University of Southern California, advised by Prof. Ulrich Neumann and Prof. Yue Wang. His research interests are in computer vision, 3D representation, generative modeling, and world models. 
\end{IEEEbiography}
\vspace{-0.6cm}

\begin{IEEEbiography}[{\includegraphics[width=1in,height=1.25in,clip,trim=0.4in 1.5in 1.2in 0in,keepaspectratio]{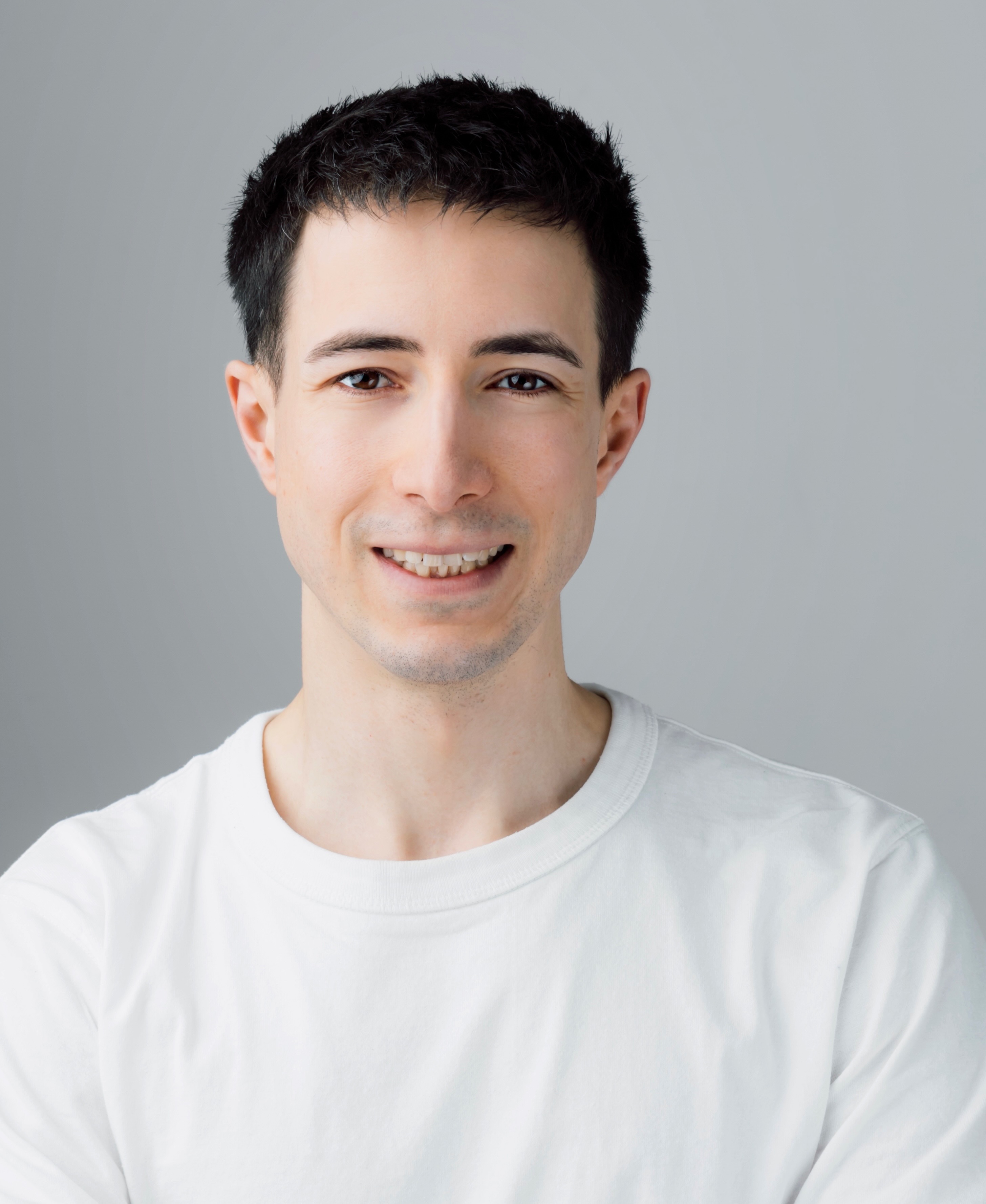}}]{Simon Guist}
received the M.Sc. degree in Applied
Computer Science from Heidelberg University, Heidelberg, Germany, in 2019. He is currently working
toward a Ph.D. degree with the Department of Empirical Inference, Max Planck Institute for Intelligent Systems, Tübingen, Germany.
His research interests lie in reinforcement learning for robotics, with emphasis on sim-to-real methods, control of tendon-/muscle-driven systems, and safe, high-speed robot control.
\end{IEEEbiography}
\vspace{-0.6cm}

\begin{IEEEbiography}[{\includegraphics[width=1in,height=1.25in,clip,keepaspectratio]
{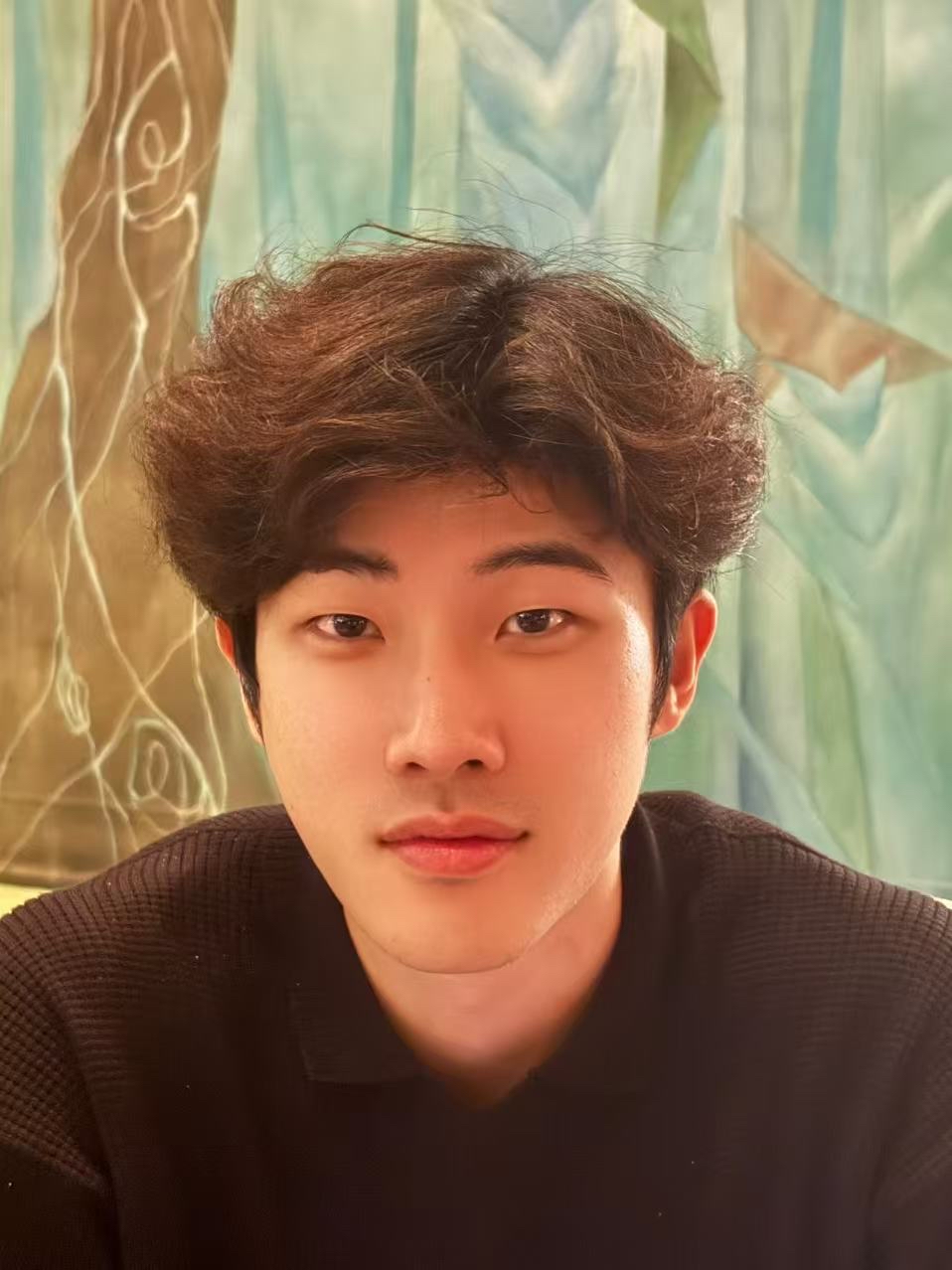}}]{Cheng Qian}
received his B.Sc. and M.Sc. degrees in Electrical Engineering and Information Technology at the Technical University of Munich. He is now a Ph.D. student in the Robot Learning Lab at Imperial College London. His primary research interest lies in developing efficient imitation learning methods for versatile and precise robotic manipulation tasks, particularly by leveraging human demonstrations.
\end{IEEEbiography}
\vspace{-0.6cm}

\begin{IEEEbiography}[{\includegraphics[width=1in,height=1.25in,clip,keepaspectratio]{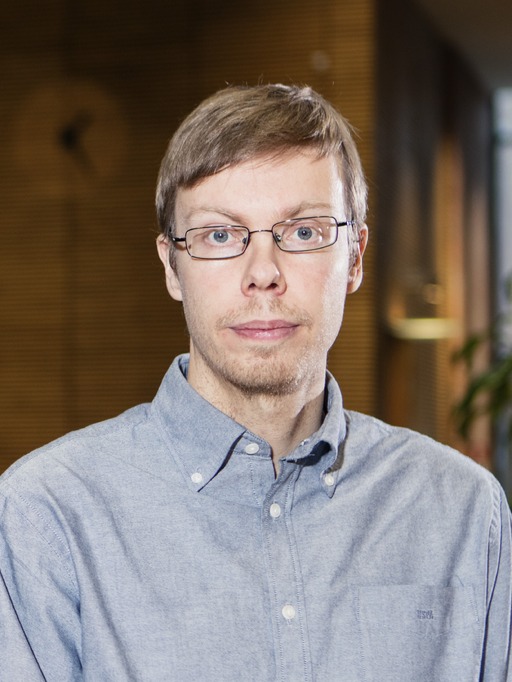}}]{Juho Kannala}
is currently an Associate Professor of Computer Vision at Aalto University and a Part-Time Professor at the University of Oulu, Finland. He received his Ph.D. degree in Signal Processing and Engineering Mathematics from the University of Oulu in 2010. From 2011 to 2014, he was a Postdoctoral Researcher funded by the Academy of Finland and served as an Academy Fellow from 2014 to 2019. His research interests include computer vision, machine learning, and biomedical image analysis. His work focuses on geometric aspects of 3D vision, recognition, and semantic image analysis. His contributions to geometric camera calibration are widely cited, and the associated software is used by researchers and practitioners worldwide.
\end{IEEEbiography}
\vspace{-0.2cm}

\begin{IEEEbiography}[{\includegraphics[width=1in,height=1.25in,clip,keepaspectratio]{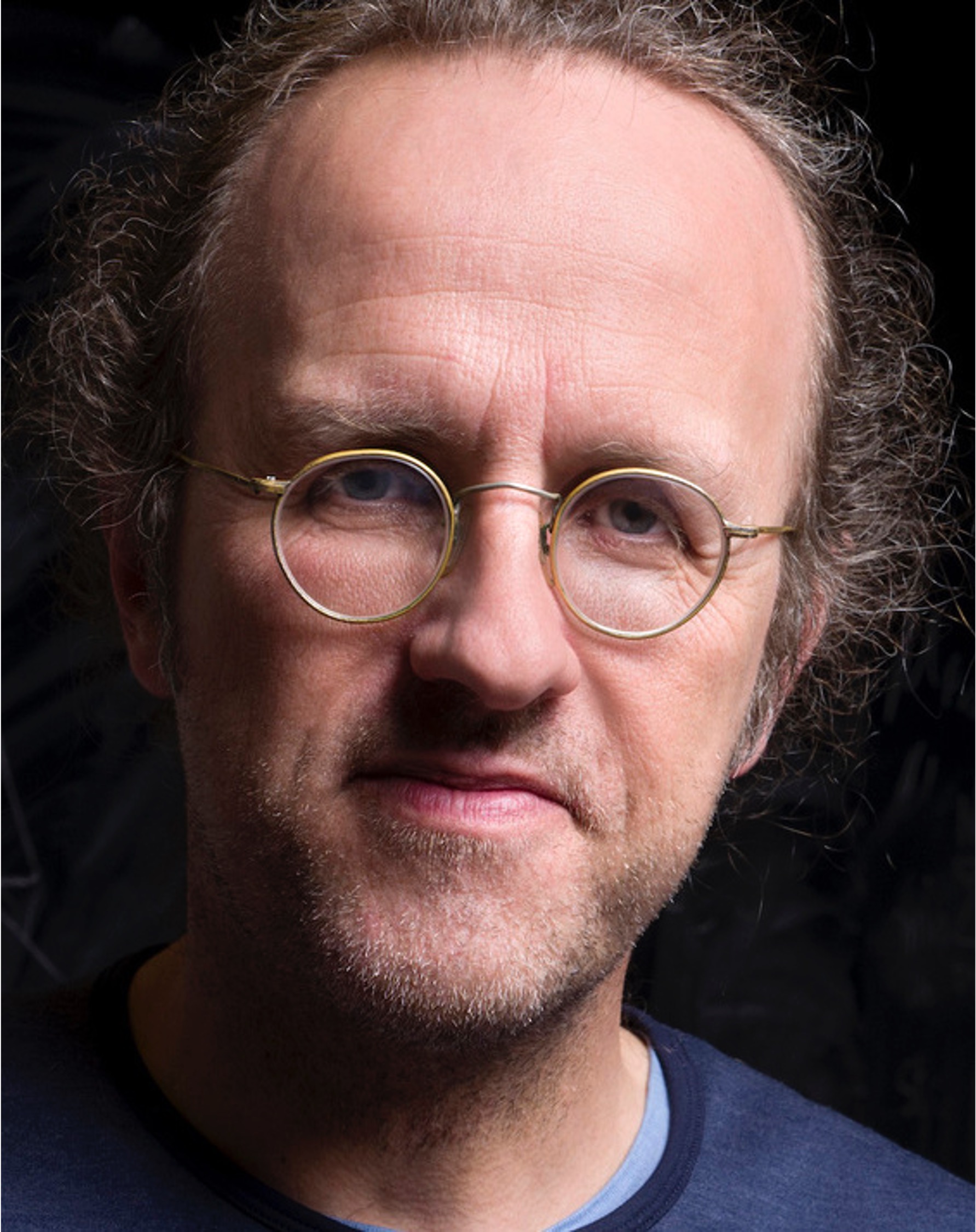}}]{Bernhard Schölkopf}
studied physics and mathematics and earned his Ph.D. in computer science in 1997, becoming a Max Planck director in 2001. He has (co-)received the Berlin-Brandenburg Academy Prize, the Royal Society Milner Award, the Leibniz Award, the BBVA Foundation Frontiers of Knowledge Award, and the ACM AAAI Allen Newell Award. He is a Fellow of the ACM and of the CIFAR Program "Learning in Machines and Brains", a member of the German Academy of Sciences, and a Professor at ETH Zurich. He helped start the MLSS series of Machine Learning Summer Schools, the Cyber Valley Initiative, the ELLIS society, and the Journal of Machine Learning Research, an early development in open access and today the field's flagship journal. In 2023, he founded the ELLIS Institute Tuebingen, and acts as its scientific director. His scientific interests are in machine learning and causal inference. He has applied his methods to a number of different fields, ranging from biomedical problems to computational photography and astronomy. 
\end{IEEEbiography}
\vspace{-0.2cm}

\begin{IEEEbiography}[{\includegraphics[width=1in,height=1.25in,clip,keepaspectratio]{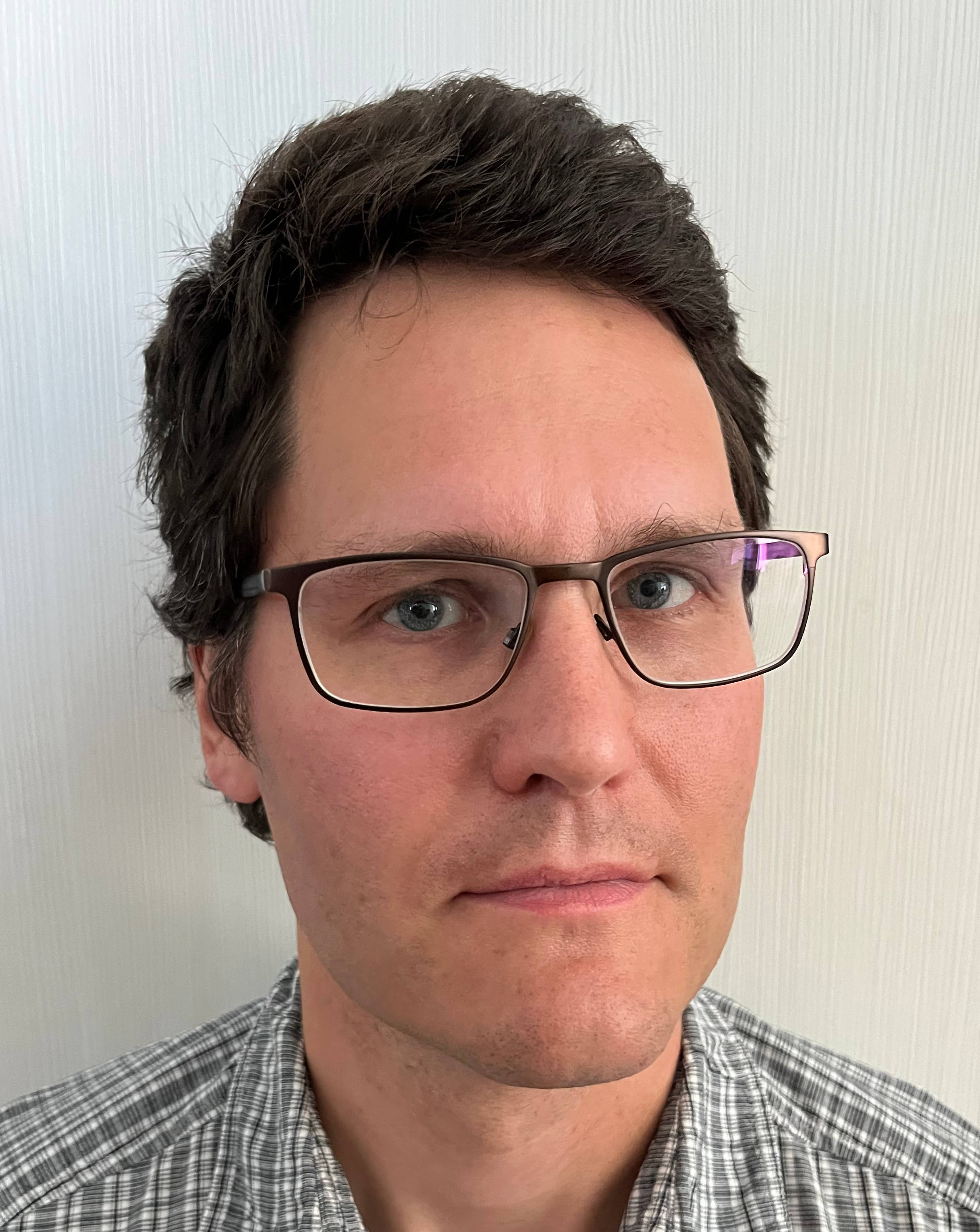}}]{Joni Pajarinen}
is an Associate Professor at Aalto University, where he leads the Aalto Robot Learning research group that develops new decision-making and control methods and applies them to novel robotic tasks. The goal of the research group is to help robots understand what they need to learn in order to perform their assigned tasks, and, thus, make robots capable of operating both on their own and in cooperation. To accomplish these goals, the research group develops computational methods and algorithms for reinforcement learning, planning under uncertainty, and decision making in multi-agent systems.
\end{IEEEbiography}
\vspace{-0.2cm}

\begin{IEEEbiography}[{\includegraphics[width=1in,height=1.25in,clip,keepaspectratio]{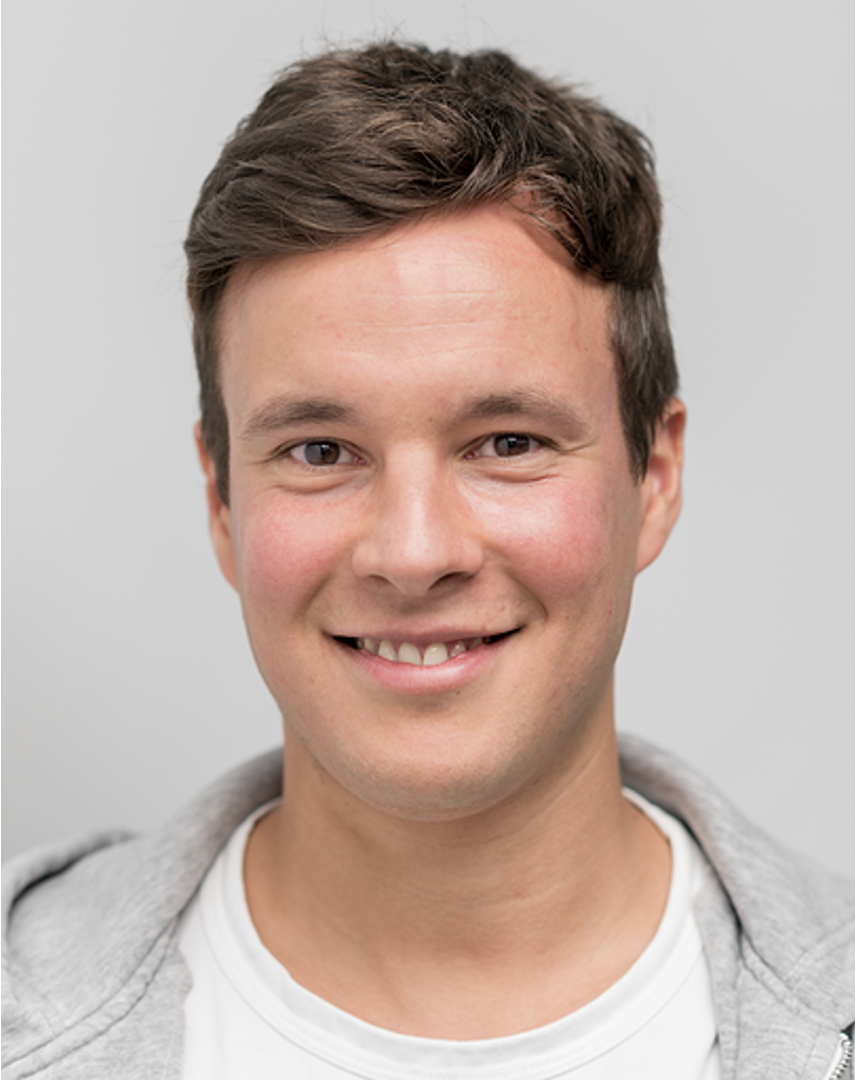}}]{Dieter Büchler} 
is an Assistant Professor in the Computing Science department at the University of Alberta, and also leads a research group in the Empirical Inference department at the Max-Planck Institute for Intelligent Systems in Tübingen, Germany. Dieter holds a Canada CIFAR AI chair and is an Alberta Machine Intelligence~(Amii) fellow. He earned a Ph.D. in Computer Science from the TU~Darmstadt under Jan Peters and Bernhard Schölkopf and performed the research at the MPI for Intelligent Systems. While pursuing his PhD, Dieter interned at X, the Moonshot Factory (formerly Google X). He holds an M.Sc. in Biomedical Engineering from Imperial College London and a B.E. in Information and Electrical Engineering from HAW Hamburg with generous support from Siemens. His mission is to achieve human performance in athletic, rapidly changing, uncertain, and high-dimensional tasks with physical robots. His research group develops learning approaches for complex systems, like soft and muscular robots, which can excel in these demanding domains. The group also studies how the robotic body influences the acquisition of robotic skills.
\end{IEEEbiography}

\vfill

{\appendices

\section{Author Contributions}

L. Chen and Y. Zhao jointly initiated and co-led the project as well as the manuscript preparation. They proposed and designed the whole approach together. L. Chen mainly worked on the multi-song agent. He implemented the final Flow Matching Transformer, performed multi-song training experiments, and conducted evaluation and ablation studies of multi-song agents. He also substantially helped with the RL experiments and curated dataset pre-processing. Y. Zhao worked mainly on the specialist agents. He implemented the OT-based fingering approach and carried out RL-related experiments. He also identified the benefits of increased dataset diversity, verified with initial experiments, and collected and curated the RP1M and RP1M++ datasets. J. Schneider contributed to Section II and substantially helped polish the manuscript. Q. Gao, S. Guist, and C. Qian contributed to writing, with C. Qian additionally helping with inverse kinematics experiments. J. Kannala, B. Schölkopf, J. Pajarinen, and D. Büchler provided project guidance and feedback; D. Büchler further contributed to manuscript polishing and project coordination.

\section{Specialist Agent and Dataset Collection Details}
\subsection{Training Details}
\paragraph{Observation Space}
Our 1144-dimensional observation space includes the proprioceptive state of dexterous robot hands and the piano as well as L-step goal states obtained from the MIDI file. In our case, we include the current goal and 10-step future goals in the observation space (L=11). At each time step, an 89-dimensional binary vector is used to represent the goal, where 88 dimensions are for key states and the last dimension is for the sustain pedal. The dimension of each component in the observation space is given in~\cref{tab:obs}. 

\begin{table}[htbp]
\caption{Observation space.}
\centering
\setlength{\extrarowheight}{0.1cm} 
\setlength\tabcolsep{8pt}
\begin{tabular}{cc}
\hline
\textbf{Observations}    & \textbf{Dim}   \\ \hline
Piano goal state         & L $\cdot$ 88   \\
Sustain goal state       & L $\cdot$ 1    \\
Piano key joints         & 88             \\
Piano sustain state      & 1              \\
Fingertip position       & 3 $\cdot$ 10   \\
Hand state               & 46             \\ \hline
\end{tabular}
\label{tab:obs}
\end{table}

\paragraph{Training Algorithm \& Hyperparameters} Although our proposed method is compatible with any reinforcement learning method, we choose the DroQ~\cite{hiraoka2021dropout} as~\cite{zakka2023robopianist} for fair comparison. DroQ is a model-free RL method, which uses Dropout and Layer normalization in the Q function to improve sample efficiency. We list the main hyperparameters used in our RL training in~\cref{tab:rl_hyperparameters}. 

% \begin{table}[htbp]
% \caption{Hyperparameters used in our RL agent.}
% \centering
% \setlength{\extrarowheight}{0.1cm} 
% \setlength\tabcolsep{8pt}
% \begin{tabular}{lc}
% \hline
% \textbf{Hyperparameter}   & \textbf{Value}   \\ \hline
% Training steps          & 8M        \\
% Episode length          & 550       \\
% Action repeat           & 1         \\
% Warm-up steps         & 5k          \\
% Buffer size             & 1M        \\
% Batch size              & 256       \\
% Update interval         & 2         \\
% Piano environment       &               \\
% ~~ Lookahead steps      & 10        \\
% ~~ Gravity compensation & True  \\
% ~~ Control timestep     & 0.05      \\
% ~~ Stretch factor       & 1.25      \\
% ~~ Trim slience         & True      \\
% Agent                   &       \\
% ~~ MLPs                 & [256, 256, 256]   \\
% ~~ Num. Q               & 2     \\
% ~~ Activation           & GeLU     \\
% ~~ Dropout Rate         & 0.01      \\
% ~~ EMA momentum         & 0.05      \\
% ~~ Discount factor      & 0.88      \\
% ~~ Learnable temperature & True     \\
% Optimization              &       \\
% ~~ Optimizer            & Adam      \\
% ~~ Learning rate        & 3e-4      \\
% ~~ $\beta_1$            & 0.9       \\
% ~~ $\beta_2$            & 0.999     \\
% ~~ eps                & 1e-8      \\
% \hline
% \end{tabular}
% \label{tab:rl_hyperparameters}
% \end{table}

\begin{table*}[htbp]
\centering
\caption{RL Training Hyperparameters and OT Fingering Reward Configuration}
\setlength{\extrarowheight}{0.15cm}
\setlength\tabcolsep{8pt}
\small
\begin{tabular}{llc}
\hline
\textbf{Category} & \textbf{Hyperparameter / Description} & \textbf{Value} \\ \hline

\multirow{7}{*}{\textbf{Training Setup}} 
  & Training steps & 8M \\
  & Episode length & 550 \\
  & Action repeat & 1 \\
  & Warm-up steps & 5k \\
  & Replay buffer size & 1M \\
  & Batch size & 256 \\
  & Update interval & 2 \\ \hline

\multirow{5}{*}{\textbf{Piano Environment}} 
  & Lookahead steps & 10 \\
  & Gravity compensation & True \\
  & Control timestep & 0.05 \\
  & Stretch factor & 1.25 \\
  & Trim silence & True \\ \hline

\multirow{5}{*}{\textbf{OT Fingering Reward Configuration}} 
  % & Reward type & Optimal Transport (Hungarian algorithm) \\
  & Distance threshold ($\delta_\text{key}$) & 0.01 \\
  & Margin ($10\times\delta_\text{key}$) & 0.1 \\
  & Tolerance function & Gaussian sigmoid \\
  % & Assignment method & \texttt{linear\_sum\_assignment} (SciPy) \\
  & Energy penalty coefficient & $5\times10^{-3}$ \\
  & Discount factor ($\gamma$) & 1.0 \\
  % & Key press reward & Tolerance-based key correctness \\ \hline

\multirow{7}{*}{\textbf{Agent Architecture}} 
  & MLP hidden layer sizes & {[}256, 256, 256{]} \\
  & Number of Q networks & 2 \\
  & Activation function & GeLU \\
  & Dropout rate & 0.01 \\
  & EMA momentum & 0.05 \\
  & Discount factor ($\gamma$) & 0.88 \\
  & Learnable temperature & True \\ \hline

\multirow{6}{*}{\textbf{Optimization}} 
  & Optimizer & Adam \\
  & Learning rate & $3\times10^{-4}$ \\
  & $\beta_1$ & 0.9 \\
  & $\beta_2$ & 0.999 \\
  & $\epsilon$ & $1\times10^{-8}$ \\
  & Weight decay & $1\times10^{-6}$ \\ \hline

\multirow{4}{*}{\textbf{Expert Agent Configuration}} 
  & Actor learning rate & $3\times10^{-4}$ \\
  & Hidden dimensions & {[}256, 256, 256{]} \\
  & Activation function & GeLU \\
  & Architecture & 3-layer MLP \\ \hline

\end{tabular}
\label{tab:rl_hyperparameters}
\end{table*}

\subsection{Computational Resources}
We train our RL agents on the LUMI cluster equipped with AMD MI250X GPUs, 64 cores AMD EPYC ``Trento" CPUs, and 64 GBs DDR4 memory. Each agent takes ~21 hours to train. The overall data collection cost is roughly 21 hours * 2089 agents = 43,869 GPU hours.

\section{Multi-Song Agent Details}

% \begin{table}[htbp]
% \caption{Hyperparameters of Diffusion Policy with UNet}
% \centering
% \setlength{\extrarowheight}{0.1cm} 
% \setlength\tabcolsep{8pt}
% \begin{tabular}{cc}
% \hline
% \textbf{Hyperparameter}  & \textbf{Value}       \\ \hline
% Batch Size Per GPU               & 2048         \\
% Optimizer                & AdamW                 \\
% Learning Rate            & 1e-4                 \\
% Weight Decay             & 1e-6                 \\
% Learning Rate Scheduler  & cosine               \\
% % Training Steps              & 1M                     \\
% Diffusion Method         & DDIM                 \\
% Number of Diffusion Iterations  & 50 \\
% Number of Inference Steps & 10 \\
% EMA Power                & 0.75                 \\
% Beta Schedule & squaredcos\_cap\_v2 \\
% U-Net Hidden Layer Sizes & {[}256, 512, 1024{]} \\
% Diffusion Step Embedding Dim. & 256             \\
% Observation Horizon      & 2                    \\
% Prediction Horizon       & 4                    \\
% Action Horizon           & 1                    \\ \hline
% \end{tabular}
% \end{table}

\begin{table*}[htbp]
\centering
\caption{Hyperparameters of the DDIM U-Net Policy}
\setlength{\extrarowheight}{0.15cm}
\setlength\tabcolsep{8pt}
\small
\begin{tabular}{llc}
\hline
\textbf{Category} & \textbf{Hyperparameter / Description} & \textbf{Value} \\ \hline

\multirow{8}{*}{\textbf{U-Net Architecture}} 
  & Diffusion step embedding dim. ($d_\text{t}$) & 256 \\
  & Downsampling channel sizes ($\text{down\_dims}$) & {[}256, 512, 1024{]} \\
  & Convolution kernel size ($k$) & 5 \\
  & GroupNorm groups ($n_\text{groups}$) & 8 \\
  & Downsampling levels & 3 \\
  & Residual blocks per level & 2 \\
  & Skip connections & Enabled \\
  & Bottleneck modules & 2 \\ \hline

\multirow{8}{*}{\textbf{DDIM Specific}} 
  & Number of training timesteps ($T_\text{train}$) & 50 \\
  & Number of inference steps ($T_\text{infer}$) & 10 \\
  & Beta schedule type & \texttt{squaredcos\_cap\_v2} \\
  & Clip denoised samples & True \\
  & Set $\alpha_T = 1$ at final step & True \\
  & Steps offset & 0 \\
  & Prediction type & \texttt{epsilon} (predict noise) \\
  & EMA power & 0.75 \\ \hline

\multirow{5}{*}{\textbf{Input / Output Dimensions}} 
  & Observation dimension ($d_\text{obs}$) & 1174 \\
  & Action dimension ($d_\text{act}$) & 39 \\
  & Prediction horizon ($H_\text{pred}$) & 4 \\
  & Action horizon ($H_\text{act}$) & 1 \\
  & Observation horizon ($H_\text{obs}$) & 2 \\ \hline

\multirow{7}{*}{\textbf{Training Hyperparameters}} 
  & Learning rate & $1\times10^{-4}$ \\
  & Weight decay & $1\times10^{-6}$ \\
  & Batch size & 10000 \\
  & Number of epochs & 2000 \\
  & Gradient accumulation steps & 1 \\
  & Optimizer & AdamW \\
  & Loss function & MSE (predicted vs. actual noise) \\ \hline

\multirow{3}{*}{\textbf{Inference Process}} 
  & Scheduler & DDIM  \\
  & Number of inference steps & 10  \\
  & EMA model usage & Enabled \\ \hline
\end{tabular}
\end{table*}

% \begin{table}[htbp]
% \caption{Hyperparameters of Flow Matching Policy with UNet}
% \centering
% \setlength{\extrarowheight}{0.1cm} 
% \setlength\tabcolsep{8pt}
% \begin{tabular}{cc}
% \hline
% \textbf{Hyperparameter}  & \textbf{Value}       \\ \hline
% Batch Size Per GPU               & 2048         \\
% Optimizer                & AdamW                 \\
% Learning Rate            & 1e-4                 \\
% Weight Decay             & 1e-6                 \\
% Learning Rate Scheduler  & cosine               \\
% Number of Training Steps & 50 \\ 
% Number of Inference Steps & 10 \\
% EMA Power                & 0.75                 \\
% Sigma & 0 \\
% U-Net Hidden Layer Sizes & {[}256, 512, 1024{]} \\
% Flow Matching Step Embedding Dim. & 256             \\
% Observation Horizon      & 2                    \\
% Prediction Horizon       & 4                    \\
% Action Horizon           & 1                    \\ \hline
% \end{tabular}
% \end{table}

\begin{table*}[htbp]
\centering
\caption{Hyperparameters of the Flow Matching U-Net Policy}
\setlength{\extrarowheight}{0.15cm}
\setlength\tabcolsep{8pt}
\small
\begin{tabular}{llc}
\hline
\textbf{Category} & \textbf{Hyperparameter / Description} & \textbf{Value} \\ \hline

\multirow{8}{*}{\textbf{U-Net Architecture}} 
  & Diffusion step embedding dim. ($d_\text{t}$) & 256 \\
  & Downsampling channel sizes ($\text{down\_dims}$) & {[}256, 512, 1024{]} \\
  & Convolution kernel size ($k$) & 5 \\
  & GroupNorm groups ($n_\text{groups}$) & 8 \\
  & Downsampling levels & 3 (256→512→1024) \\
  & Residual blocks per level & 2 \\
  & Skip connections & Enabled \\
  & Bottleneck modules & 2 \\ \hline

\multirow{3}{*}{\textbf{Flow Matching Specific}} 
  & Noise level ($\sigma$) & 0.0 \\
  & EMA power & 0.75 \\
  & Loss function & MSE (predicted vs. target flow) \\ \hline

\multirow{5}{*}{\textbf{Input / Output Dimensions}} 
  & Observation dimension ($d_\text{obs}$) & 1174 \\
  & Action dimension ($d_\text{act}$) & 39 \\
  & Prediction horizon ($H_\text{pred}$) & 4 \\
  & Action horizon ($H_\text{act}$) & 1 \\
  & Observation horizon ($H_\text{obs}$) & 2 \\ \hline

\multirow{7}{*}{\textbf{Training Hyperparameters}} 
  & Learning rate & $1\times10^{-4}$ \\
  & Learning Rate Scheduler  & Cosine  \\
  & Weight decay & $1\times10^{-6}$ \\
  & Batch size & 10000 \\
  & Number of epochs & 2000 \\
  & Gradient accumulation steps & 1 \\
  & Optimizer & AdamW \\ \hline

\multirow{3}{*}{\textbf{Inference}} 
  & Integration method & Euler (step size $dt = 1/10$) \\
  & Number of inference steps & 10 \\
  & EMA model usage & Enabled \\ \hline
\end{tabular}
\end{table*}

\begin{table*}[htbp]
\centering
\caption{Model Architecture and Training Configuration of the Flow Matching Transformer Policy}
\setlength{\extrarowheight}{0.15cm}
\setlength\tabcolsep{8pt}
\small
\begin{tabular}{llc}
\hline
\textbf{Category} & \textbf{Hyperparameter} & \textbf{Value} \\ \hline
\multirow{9}{*}{\textbf{Model Architecture}} 
  & Number of transformer layers ($n_\text{layer}$) & 12 \\
  & Number of attention heads ($n_\text{head}$) & 12 \\
  & Embedding dimension ($n_\text{emb}$) & 768 \\
  & Embedding dropout ($p_\text{drop\_emb}$) & 0.0 \\
  & Attention dropout ($p_\text{drop\_attn}$) & 0.1 \\
  & Causal attention & False (bidirectional) \\
  & Time conditioning ($\text{time\_as\_cond}$) & True \\
  & Observation conditioning ($\text{obs\_as\_cond}$) & True \\
  & Conditioning layers ($n_\text{cond\_layers}$) & 0 \\ \hline

\multirow{5}{*}{\textbf{Flow Matching Specific}} 
  & Noise level ($\sigma$) & 0.0 \\
  & Number of projections & 100 \\
  & Sliced OT loss & Disabled \\
  & Antithetic sampling & Disabled \\
  & EMA power & 0.75 \\ \hline

\multirow{5}{*}{\textbf{Input / Output Dimensions}} 
  & Observation dimension ($d_\text{obs}$) & 1144 \\
  & Action dimension ($d_\text{act}$) & 39 \\
  & Prediction horizon ($H_\text{pred}$) & 4 \\
  & Action horizon ($H_\text{act}$) & 1 \\
  & Observation horizon ($H_\text{obs}$) & 2 \\ \hline

\multirow{7}{*}{\textbf{Training Hyperparameters}} 
  & Learning rate & $1\times10^{-4}$ \\
  & Weight decay & $1\times10^{-3}$ \\
  & Batch size & 10000 \\
  & Number of epochs & 2000 \\
  & Gradient accumulation steps & 1 \\
  & Optimizer & AdamW \\
  & Mixed precision & bfloat16 \\ \hline

\multirow{5}{*}{\textbf{Optimization \& Stability}} 
  & Gradient clipping ($\lVert g\rVert_\infty$) & 0.1 \\
  & LR scheduler type & Cosine (with warmup) \\
  & LR warmup steps & 1000 \\
  & Minimum LR & $1\times10^{-6}$ \\
  & EMA stability & Enabled \\ \hline

\multirow{2}{*}{\textbf{Inference}} 
  & Number of inference steps & 10 \\
  & Integration method & Euler (step size $dt = 1/10$) \\ \hline
\end{tabular}
\end{table*}

\subsection{Training and Evaluation}
\label{appendix:mt_train-eval}
We train the policies with different sizes of expert data: 12, 150, 300, 500, 700, and 900 songs, respectively. Subsequently, we assess the trained policies using two distinct categories of musical pieces. The first category, in-distribution songs, includes pieces that are part of the training datasets. Evaluating with in-distribution songs tests the multitasking abilities of the policies and checks if a policy can accurately recall the songs on which it was trained. The second group of songs for evaluation are out-of-distribution songs: those music pieces do not overlap with the training songs. The selected songs contain diverse motions and long horizons, making them challenging to play. This out-of-distribution evaluation measures the zero-shot generalization capabilities of the policies. Analogous to an experienced human pianist who can play new pieces at first sight, we aim to determine if it is feasible to develop a generalist agent capable of playing the piano under various conditions.

% Additionally, our framework is designed with flexibility in mind, allowing users to select songs not included in our dataset for either training data collection or evaluation. Furthermore, users have the option to assess their policies on specific segments of a song rather than the entire piece.

\begin{table}[htbp]
\caption{Out-of-distribution songs (Part 1)}
\centering
\setlength{\extrarowheight}{0.1cm}
\setlength\tabcolsep{8pt}
\begin{tabular}{l}
\hline
RoboPianist-GP-CarbajoVictorYellowSaraband-v0 \\
RoboPianist-GP-OestenTheodoreDerJungfrauGebetAmVerlobun-v0 \\
RoboPianist-GP-OestenTheodoreMaiblumchenOp61-v0 \\
RoboPianist-GP-MoszkowskiMoritz3MomentsMusicauxOp7-v0 \\
RoboPianist-GP-MoszkowskiMoritz6MorceauxOp31-v0 \\
RoboPianist-GP-MoszkowskiMoritzGondolieraOp41-v0 \\
RoboPianist-GP-CanteloubeJosephDansesRoumaines-v0 \\
RoboPianist-GP-StithDavidJosephPavane-v0 \\
RoboPianist-GP-HinloopenTooskeSpanishSerenade-v0 \\
RoboPianist-GP-ChavagnatEdouardEtoileDuMatinOp201-v0 \\
RoboPianist-GP-GottschalkLouisMoreauOrfaOp71-v0 \\
RoboPianist-GP-GottschalkLouisMoreauPensiveOp68-v0 \\
RoboPianist-GP-ShcherbachyovNikolayScherzoCapriceOp17-v0 \\
RoboPianist-GP-WielhorskiJozef2ValsesOp21-v0 \\
RoboPianist-GP-WielhorskiJozef4BagatellesOp47-v0 \\
RoboPianist-GP-WieniawskiJozefBalladeOp31-v0 \\
RoboPianist-GP-WieniawskiJozefFantaisieEtFugueOp25-v0 \\
RoboPianist-GP-NowakowskiJozefFantaisieSurLoperaHalkaOp-v0 \\
RoboPianist-GP-WieniawskiJozefNocturneOp37-v0 \\
RoboPianist-GP-WieniawskiJozefReverieOp45-v0 \\
RoboPianist-GP-WieniawskiJozefValseCapriceOp46-v0 \\
RoboPianist-GP-WieniawskiJozefValseDeConcertNo1Op3-v0 \\
RoboPianist-GP-MagnenatStephaneVariationsDautomne-v0 \\
RoboPianist-GP-BergerRodolpheDansLesFleurs-v0 \\
RoboPianist-GP-BergerRodolpheLoinDuPays-v0 \\
RoboPianist-GP-RiottePhilippJakobRondeauOp1-v0 \\
RoboPianist-GP-ZintlFrankSpringar-v0 \\
RoboPianist-GP-LullyJeanBaptiste4Buhnenwerke-v0 \\
RoboPianist-GP-LourieArthur2Compositions-v0 \\
RoboPianist-GP-HintonArthurASummerPilgrimage-v0 \\
RoboPianist-GP-BendixVictorIntermezzo-v0 \\
RoboPianist-GP-TinelJefDaarWaarMooieBloemkensBloeien-v0 \\
RoboPianist-GP-MinotAdolfDramaticAgitato-v0 \\
RoboPianist-GP-LambJosephFrancisEthiopiaRag-v0 \\
RoboPianist-GP-HitzFranzATraversBoisOp151-v0 \\
RoboPianist-GP-ButlerLeonardAsterisks-v0 \\
RoboPianist-GP-ButlerLeonardAtNightfall-v0 \\
RoboPianist-GP-StraussEduardConAmoreOp60-v0 \\
RoboPianist-GP-KugeleRichardEinsamesRosleinOp17-v0 \\
RoboPianist-GP-ButlerLeonardTheSongOfTheFountain-v0 \\
RoboPianist-GP-WagnerRichardZuricherVielliebchenWalzerW-v0 \\
RoboPianist-GP-FerroudPierreOctavePreludeEtForlane-v0 \\
RoboPianist-GP-CarrascoAlfredoRomanzaSinPalabras-v0 \\
RoboPianist-GP-SatieErik4Preludes-v0 \\
RoboPianist-GP-PfeifferGeorgesJean25EtudesOp70-v0 \\
RoboPianist-GP-ProkofievSergey3PiecesFromCinderellaOp95-v0 \\
RoboPianist-GP-KowalskiHenriMenuetDeMarieLesczynskaOp23-v0 \\
RoboPianist-GP-WeberCarlMariaVon12AllemandesOp4-v0 \\
RoboPianist-GP-ChopinFredericMarcheFunebreOp72No2-v0 \\
RoboPianist-GP-ChopinFredericMazurkaInDMajorB31-v0 \\
RoboPianist-GP-ChopinFredericNocturneOubliee-v0 \\ \hline
\end{tabular}
\end{table}

\begin{table}[htbp]
\caption{Out-of-distribution songs (Part 2)}
\centering
\setlength{\extrarowheight}{0.1cm}
\setlength\tabcolsep{8pt}
\begin{tabular}{l}
\hline
RoboPianist-GP-ChopinFredericValseMelancoliqueInFSharpM-v0 \\
RoboPianist-GP-ChopinFredericWaltzInEMinorB56-v0 \\
RoboPianist-GP-LichnerHeinrichFenellaOp48-v0 \\
RoboPianist-GP-HofmannHeinrichGavotte-v0 \\
RoboPianist-GP-LichnerHeinrichHerzenswunscheOp5-v0 \\
RoboPianist-GP-HofmannHeinrichNachklangeBook2Op37-v0 \\
RoboPianist-GP-KuhlauFriedrichPianoSonatinaInDMajorOp55-v0 \\
RoboPianist-GP-ZwyssigAlberichSchweizerpsalm-v0 \\
RoboPianist-GP-HofmannHeinrichStimmungsbilderOp88-v0 \\
RoboPianist-GP-ZelterCarlFriedrichKeyboardSonataInFMajo-v0 \\
RoboPianist-GP-MesquitaHenriqueAlvesDeLaBresilienne-v0 \\
RoboPianist-GP-MesquitaHenriqueAlvesDeLaCoquette-v0 \\
RoboPianist-GP-MarcozziMaurizioInvenzioneA2VociNo18-v0 \\
RoboPianist-GP-NeustedtCharlesMenuetSentimental-v0 \\
RoboPianist-GP-KoechlinCharlesPaysagesEtMarinesOp63-v0 \\
RoboPianist-GP-SchreinerHermannLBeauregardsCharlestonQu-v0 \\
RoboPianist-GP-SchreinerHermannLStonewallJacksonsGrandM-v0 \\
RoboPianist-GP-KurpinskiKarolChwilaSnuOkropnego-v0 \\
RoboPianist-GP-RopartzGuyCroquisDete-v0 \\
RoboPianist-GP-VonTilzerHarryCornfieldCapers-v0 \\
RoboPianist-GP-VonTilzerHarryGhostOfTheGoblinMan-v0 \\
RoboPianist-GP-VonTilzerHarryILoveIt-v0 \\
RoboPianist-GP-LeonardRobertAbendfriedenOp153-v0 \\
RoboPianist-GP-KussnerAlbertMemoriesOp6-v0 \\
RoboPianist-GP-AuricGeorges3Pastorales-v0 \\
RoboPianist-GP-WeissSylviusLeopoldLuteSonataInDMinorWei-v0 \\
RoboPianist-GP-WidorCharlesMarie6DuosOp6-v0 \\
RoboPianist-GP-WidorCharlesMarieCarnavalOp61-v0 \\
RoboPianist-GP-WidorCharlesMarieScherzoValseOp5-v0 \\
RoboPianist-GP-WidorCharlesMarieValsesOp33-v0 \\
RoboPianist-GP-MayerCharlesNocturneOp81-v0 \\
RoboPianist-GP-MorleyThomasSingingAlone-v0 \\
RoboPianist-GP-EgghardJulesUneRoseDesMontagnesOp85-v0 \\
RoboPianist-GP-KopylovAleksandrMazurkaOp8-v0 \\
RoboPianist-GP-KopylovAleksandrPolkaDeSalonOp16-v0 \\
RoboPianist-GP-TostiFrancescoPaoloPepita-v0 \\
RoboPianist-GP-CileaFrancescoSerenata-v0 \\
RoboPianist-GP-WachsPaulRoseDautomne-v0 \\
RoboPianist-GP-OlsenOleValseIntime-v0 \\
RoboPianist-GP-MattheyUlissePensieroOstinato-v0 \\
RoboPianist-GP-ZoellerErnestFantaisieOnMignon-v0 \\
RoboPianist-GP-VincentAugusteLaDeclarationOp28-v0 \\
RoboPianist-GP-SindingChristianFantaisiesOp118-v0 \\
RoboPianist-GP-TeilmanChristianITroldheimen-v0 \\
RoboPianist-GP-TeilmanChristianImprovistionVedHarpenOp1-v0 \\
RoboPianist-GP-TeilmanChristianSouvenirDeWieniawskiOp16-v0 \\
RoboPianist-GP-TeilmanChristianSvundneTider-v0 \\
RoboPianist-GP-LecuonaErnestoMazurkaGlissando-v0 \\ \hline
\end{tabular}
\end{table}

\begin{table}[htbp]
\caption{In-distribution songs}
\centering
\setlength{\extrarowheight}{0.1cm}
\setlength\tabcolsep{8pt}
\begin{tabular}{l}
\hline
RoboPianist-etude-12-FrenchSuiteNo1Allemande-v0 \\
RoboPianist-etude-12-FrenchSuiteNo5Sarabande-v0 \\
RoboPianist-etude-12-PianoSonataD8451StMov-v0 \\
RoboPianist-etude-12-PartitaNo26-v0 \\
RoboPianist-etude-12-WaltzOp64No1-v0 \\
RoboPianist-etude-12-BagatelleOp3No4-v0 \\
RoboPianist-etude-12-KreislerianaOp16No8-v0 \\
RoboPianist-etude-12-FrenchSuiteNo5Gavotte-v0 \\
RoboPianist-etude-12-PianoSonataNo232NdMov-v0 \\
RoboPianist-etude-12-GolliwoggsCakewalk-v0 \\
RoboPianist-etude-12-PianoSonataNo21StMov-v0 \\
RoboPianist-etude-12-PianoSonataK279InCMajor1StMov-v0 \\ \hline
\end{tabular}
\end{table}

}

\end{document}